\documentclass[11pt]{article}

\usepackage[final]{acl}

\usepackage{times}
\usepackage{latexsym}
\usepackage{amsmath}
\usepackage[dvipsnames]{xcolor}

\usepackage[T1]{fontenc}

\usepackage[utf8]{inputenc}

\usepackage{microtype}

\usepackage{inconsolata}

\usepackage{graphicx}

\usepackage{listings}
\usepackage{subcaption}
\usepackage{booktabs}
\usepackage{makecell}
\usepackage{multirow}
\usepackage{enumitem}

\newcommand{\xhdr}[1]{\vspace{1mm}\noindent{{\bf #1}}}

\newlength{\ctabmodel}
\newlength{\ctabstage}
\newlength{\ctabnum}
\newcommand{\cmod}[1]{\makebox[\ctabmodel][l]{#1}}
\newcommand{\cstg}[1]{\makebox[\ctabstage][l]{#1}}
\newcommand{\cnum}[1]{\makebox[\ctabnum][c]{#1}}
\newcommand{\dashsep}[1]{\noalign{\vskip1.5pt}%
  \multicolumn{#1}{@{}l@{}}{\rlap{\makebox[\dimexpr\ctabmodel+\ctabstage+5\ctabnum+14\tabcolsep\relax][l]{%
    \leaders\hbox to 4pt{\hss\rule[0.5ex]{2pt}{0.4pt}\hss}\hfill\kern0pt}}}\\[1.5pt]}

\definecolor{mygreen}{RGB}{101, 179, 113}
\definecolor{ntpblue}{RGB}{72, 102, 207}
\definecolor{randpurp}{RGB}{76, 123, 121}
\definecolor{auggreen}{RGB}{154, 122, 77}

\title{Learning Concepts, Not Tokens:\\Self-Supervised Semantic Alignment for Language Models}

\author{Christine Zhang\textsuperscript{1}, Dan Jurafsky\textsuperscript{1}, Chen Shani\textsuperscript{2, 1} \\
  \textsuperscript{1}Stanford University, \textsuperscript{2}Tel Aviv University \\
  \texttt{chrzhang@stanford.edu, jurafsky@stanford.edu, cshani@tauex.tau.ac.il}}

\begin{document}
\maketitle
\begin{abstract}
The next-token prediction (NTP) objective trains language models to predict a single token at each step, even though many continuations can express the same meaning. For example, in the sentence ``this sticker can be \textit{placed} here'', \textit{positioned}, \textit{attached}, or \textit{put} are all plausible alternatives. While standard NTP training treats these alternatives as mutually exclusive targets, we explore a self-supervised framework that encourages models to predict concepts, approximated as sets of semantically equivalent tokens. Models trained with this concept supervision align better with human similarity judgments, improve classification, clustering, and reranking performance, and achieve comparable or stronger downstream reasoning. These gains come with lower perplexity on semantically meaningful words (Section~\ref{sec:data}) and only minimal increases in global perplexity, suggesting that concepts enhance semantic alignment while preserving language modeling quality.\footnote{Our code is available at \url{https://github.com/christine-zhang1/learning-concepts}.}

\end{abstract}

\section{Introduction}

Large language models (LLMs) can generate coherent text, answer questions, and perform complex reasoning, yet most are still trained with a simple objective: next token prediction (NTP). While effective in practice, NTP emphasizes surface-level token patterns, treating different wordings as distinct even when they express the same meaning \cite{michaelov2025not}.

In contrast, humans reason in terms of concepts that can be expressed through different words and contexts. For example, \textit{dad} and \textit{father} refer to the same entity, while words like \textit{dog} and \textit{pet} can evoke the same concept in context. As a result, NTP may penalize valid alternatives despite their semantic equivalence, diverging from human conceptual processing \citep{bachmann2024pitfalls, liu2025predict, shani2026tokens}.

Here, we modify the training objective to \textbf{predict concepts rather than surface-level tokens}, where we approximate concepts as sets of contextual synonyms. We introduce a self-contained training pipeline that leverages the model's own knowledge to make its representations more concept-aware.

Our concept-tuned models, trained on data with short contexts, show \textbf{stronger alignment with human semantic intuition}, improved performance on short-context classification, clustering, and reranking, and exhibit comparable or stronger downstream reasoning abilities. Moreover, despite not being trained to predict exact tokens, concept-trained models achieve lower perplexity on content words (defined in Section~\ref{sec:data}) than all baselines, with only slightly higher global perplexity over all tokens.

\section{Related Work}

\begin{figure*}[h]
    \centering   
    \includegraphics[width=\linewidth]{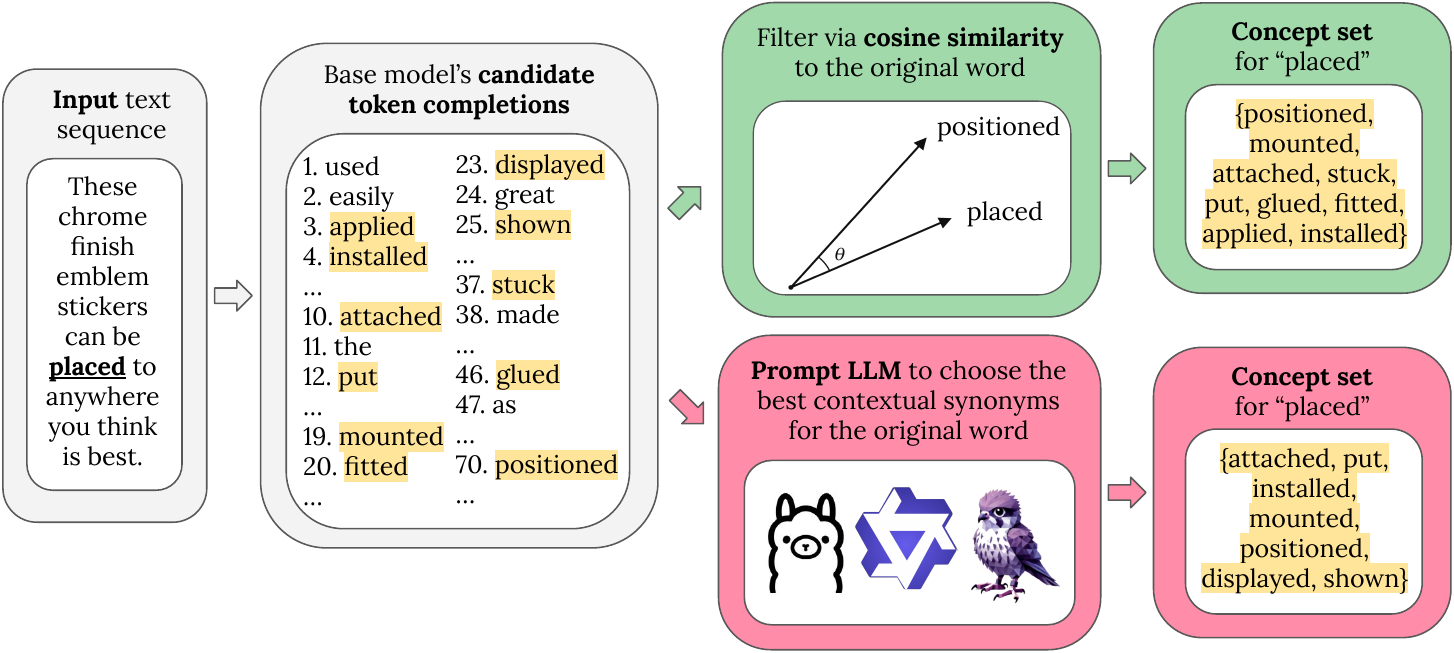}
    \caption{Our process for concept dataset generation, illustrated on one word for this sequence. We pass each sequence into a decoder model, and for each noun, verb, or adjective token that comprises an entire word, we extract the top 100 NTP completions. Then, we select the best contextual synonyms for those words using two different methods. In \textcolor{mygreen}{method 1}, we compare the cosine similarities between the original word's embedding and the embedding of each candidate token, and we keep the candidate tokens that have a cosine similarity of at least 0.75. In \textcolor{WildStrawberry}{method 2}, we prompt the model's corresponding instruct-tuned model to choose the best synonyms from the candidate tokens. These synonyms form the concept set for the original word in the sequence.}
    \label{fig:pipeline}
\end{figure*}

NTP's reliance on exact token matching can penalize correct concepts, distort token rankings, and produce undesirable behaviors \cite{holtzman2021surface, shani2023towards, iyer2026beyond, liu2026profit}. \citet{bender-koller-2020-climbing} argue that this surface-level focus limits semantic understanding, a claim supported by failures of NTP and teacher forcing on tasks requiring entailment, planning, or multiple valid continuations \cite{bachmann2024pitfalls, merrill2024can}. Together, these findings suggest that fluent generation does not necessarily imply semantic understanding. 

A broad line of work has sought to relax the assumption that learning should assign all credit to a single observed output. Early approaches distributed probability mass over alternative tokens through label and loss smoothing, or over nearby sequences according to task-specific similarity or reward functions \cite{elbayad-etal-2018-token, gao-etal-2020-towards, lukasik-etal-2020-semantic, norouzi2016reward}. Subsequent work extended this idea to semantically similar or equivalent sequences, either through training objectives \cite{lukasik-etal-2020-semantic, khayrallah-sedoc-2020-smrt} or inference-time criteria \cite{zhou-etal-2019-bert, arefyev-etal-2020-always, xiong2026superthoughts}. A newer line of concept-related work uses continuous, soft tokens as intermediate computation steps \cite{zhang2026soft, butt2026soft, zheng2025soft, zhao2026learning, barrault2024large}. While promising, all of these approaches emerged across different tasks, motivations, and terminology, making it difficult to recognize them as part of a coherent line of work.

Recent work has explored alternatives to standard \textit{pretraining}, including multi-token prediction \cite{deepseekai2025deepseekv3technicalreport, ahn2025efficient, chen2025improving}, plan-then-generate frameworks \cite{mahajan2025beyond, wei2025plangenllms}, latent reasoning \cite{barrault2024large, tack2026llm}, continuous generation \cite{li2022diffusion, shabalin2025tencdm}, and non-transformer architectures \cite{gu2024mamba, huang2026llmjepa}. Other approaches instead address NTP's limitations during \textit{post-training} through reinforcement learning from human feedback, direct preference optimization, instruction tuning, contrastive learning, and multitask finetuning \cite{ouyang2022training, wu2023fine, rafailov2023direct}. For example, semantic augmentations improve generalization \cite{yan2024contrastive}, while lexical supervision improves word sense disambiguation \cite{levine2020sensebert}.

The most closely related work incorporates concept-level information into the training signal through next-synonym and hypernym prediction \cite{iyer2026beyond}. Although effective, these methods are not self-contained, sometimes relying on external resources such as WordNet to define concept sets. In contrast, we introduce a method to derive concept structure directly from the embeddings of the base model that we post-train. Moreover, \citet{iyer2026beyond} apply concept training only to nouns, while we extend the approach to include verbs and adjectives to capture more of the semantically meaningful words in each sequence.

\section{Training Methods}

We first list the base models used (Section \ref{sec:base_models}). We then describe both the process of enriching standard data with concept signal (Section \ref{sec:data}; Figure \ref{fig:pipeline}), and modifying standard NTP objective function to be concept-aware (Section \ref{sec:objective_func}; Figure \ref{fig:objective}).  

\subsection{Base Models}
\label{sec:base_models}
Throughout the paper, we explore 3 model families and 3 model sizes within each family. We use: 


\begin{itemize}[noitemsep,topsep=2pt,parsep=0pt,partopsep=0pt,leftmargin=*]
    \item Llama 3.2 1B, Llama 3.2 3B, Llama 3.1 8B~\cite{grattafiori2024llama}
    \item Qwen 3 1.7B, Qwen 3 4B, Qwen 3 8B~\cite{qwen3technicalreport}
    \item Falcon 3 1B, Falcon 3 3B, Falcon 3 7B~\cite{Falcon3}
\end{itemize}

\subsection{Concept Dataset Creation}
\label{sec:data}

\begin{figure*}[t]
    \centering
    \includegraphics[width=0.98\linewidth]{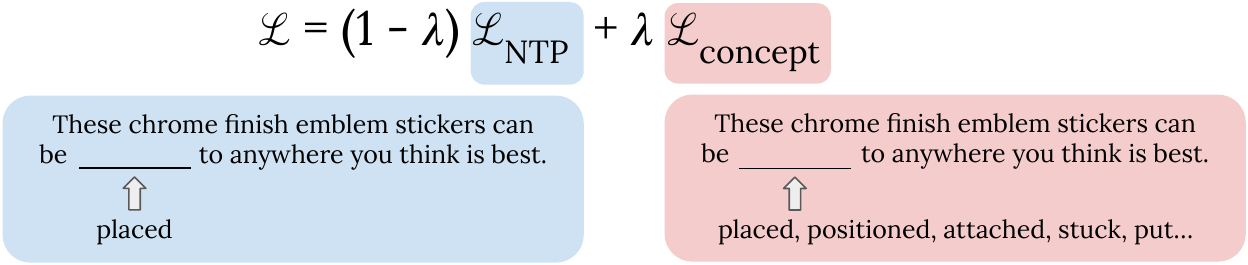}
    \caption{Our loss function incorporating both standard NTP and a concept term, balanced via a concept weight hyperparameter $\lambda$. Setting $\lambda=0$ is pure NTP training, and increasing $\lambda$ makes the models more concept-oriented.}
    \label{fig:objective}
\end{figure*}

We use two standard datasets: AllenAI's C4 \cite{raffel2020exploring} and OpenWebText \cite{Gokaslan2019OpenWeb}, an open-source replication of WebText. From each corpus, we sample 10,000 English text sequences and split them into training, validation, and test sets using an 80/10/10 ratio. We follow Sentence Transformers \cite{reimers2019sentence} and truncate sequences to 256 tokens. Models are thus trained on $10{,}000 \times 0.8 \times 256 = 2{,}048{,}000$ tokens from either C4 or OpenWebText.

We use two methods to generate concept datasets: one operates within each model, and the other uses its instruction-tuned variant. Since both are model-dependent, we generate a separate dataset for each base model in Section~\ref{sec:base_models}.

Both methods first identify token positions for concept-level supervision. We target tokens that (i) correspond to complete words rather than subwords and (ii) are nouns, verbs, or adjectives, ensuring that meaningful semantic alternatives exist.


To identify complete words, we retain tokens that begin with a space and are followed by either another space-prefixed token or punctuation, selecting standalone words rather than subword fragments. This excludes the first word in each sequence, since lacking preceding context prevents reliable identification of contextual synonyms.


We use spaCy POS tagging and verify alignment with the tokenizer. On average, complete-word single-token nouns, verbs, and adjectives (\textit{content words}) comprise 24\% of tokens per sequence.

For each content-word position, we generate the top 100 next-token predictions as candidates that express the same concept. Our two methods then differ in how they filter this pool into a final set of contextually interchangeable tokens (Figure~\ref{fig:pipeline}).

\xhdr{\textcolor{mygreen}{Method 1: Embeddings. }}We compute the cosine similarity between each candidate's contextual embedding and that of the original token. We discard candidates whose spaCy part-of-speech tag does not match the original. From the remaining pool, we retain candidates with cosine similarity at least 0.75, capped at 10 per content word. The resulting set serves as a discrete approximation of the concept associated with that token in its context.

\xhdr{\textcolor{WildStrawberry}{Method 2: Prompting. }}We prompt the model's corresponding instruction-tuned variant to select which of the 100 candidates can replace the original token without changing the sentence's meaning (see Appendix~\ref{appendix:instruct_prompt} for the prompt used).

We include details on the size of concept sets for each model, dataset, and concept generation method in Tables~\ref{tab:content_words} and~\ref{tab:concept_sizes} in Appendix~\ref{appendix:concept_stats}. Concept sets have been spot-checked by humans for quality.

\subsection{Concept Training Objective}
\label{sec:objective_func}

We introduce a \textit{concept-level loss} that rewards semantically equivalent completions, rather than a single surface form. 

Let $S$ denote an input sentence (e.g., ``this sticker can be placed''), and let $T$ be the original content token (``placed''). We define $T^\ast$ as the set of contextual synonyms for $T$, including $T$ itself: \{\textit{placed, attached, put, installed, ...}\}. Instead of optimizing the likelihood of predicting the exact token $T$, the concept objective rewards predicting \textit{any} token belonging to the conceptual equivalence class $T^\ast$:
\[
\mathcal{L}_{\text{concept}}(T^\ast \mid S)
= -\log
\sum_{n=1}^{|T^\ast|}
 p(t_n \in T^\ast \mid S, \Theta)
\]

\noindent where $\Theta$ denotes the model parameters and $p(t_n \in T^\ast \mid S, \Theta)$ is the probability assigned by the model to the $n$-th token in the contextual synonym set $T^\ast$. This objective \textbf{encourages the model to distribute probability mass across multiple realizations of the same concept}, rather than concentrating exclusively on a single lexical form.

\subsection{Combined Training Loss}

We apply standard NTP loss to non-content words, and for content words combine NTP and concept loss via a weighted linear interpolation (Figure~\ref{fig:objective}):
\[
\mathcal{L}_{\text{total}}
= (1 - \lambda)\,\mathcal{L}_{\text{NTP}}
+ \lambda\,\mathcal{L}_{\text{concept}}
\]
where $\lambda \in [0,1]$ controls the strength of concept-level supervision. Setting $\lambda = 0$ is equivalent to standard NTP training, while larger values place increasing emphasis on concept prediction.

\subsection{Models}
\label{sec:models}
We post-train our nine base models using concept loss weights $\lambda \in \{0.25, 0.5, 0.75, 1.0\}$. We train separately on C4 and OpenWebText, and separately on concept datasets produced by each of our two generation methods, yielding 144 total combinations (9 base models $\times$ 4 concept loss weights $\times$ 2 datasets $\times$ 2 concept dataset generation methods).

\subsection{Baselines}

\label{sec:baselines}
We compare against four baselines, each designed to isolate a specific potential source of gains.

\paragraph{\textcolor{black}{Base model.}} The base model with no post-training, to measure the effect of post-training.

\paragraph{\textcolor{ntpblue}{NTP.}} We post-train models with only NTP loss ($\lambda=0$) on the same data, to measure the contribution of concept signal.

\paragraph{\textcolor{randpurp}{Randomized synonym sets.}} To verify that the observed gains stem from semantically coherent synonym sets rather than from lexical variation acting merely as a regularizer, we train models in which each true synonym set is replaced with a random set of the same size, sampled from the global pool of training-set synonyms. We use $\lambda \in {0.25, 0.5, 0.75, 1.0}$, since the $\lambda = 0$ condition is already covered by the previous baseline.

\paragraph{\textcolor{auggreen}{Data-augmented NTP.}} To verify that gains stem from the concept objective itself rather than from increased exposure to semantically similar words, we train NTP models on a concept-augmented dataset.  For each original sequence, we generate four additional variants in which every content word is replaced by a sample from its corresponding concept set. When a concept set contains at least four synonyms, samples are drawn without replacement; otherwise, sampling is performed with replacement (including the original word). We train two variants: one matched to our main models in total training steps (controlling for compute), and another matched in epochs (controlling for unique data exposure), the latter requiring five times as many training steps. This data-augmented baseline also mirrors the best-performing configuration from \citet{iyer2026beyond}, allowing a direct comparison against the strongest prior approach.

\subsection{Training Details}
All models are trained using QLoRa-4bit \cite{dettmers2023qlora} with a learning rate of $7 \times 10^{-5}$ and batch size of 2 for $5$ epochs. We limit the contextual synonym set size to a maximum of $10$ excluding the content word itself, and we use a probability mass threshold of $0.6$. The models take about 0.5-2 hours to train on an NVIDIA A6000 GPU depending on model size and dataset.

To prevent the concept objective from dominating NTP, we introduce a probability mass threshold on the contextual synonym set. If the cumulative probability assigned by the model to tokens in $T^\ast$ exceeds a certain threshold (we used $0.6$), we set $\mathcal{L}_{\text{concept}} = 0$ for that training example. This ensures that once the model sufficiently captures the concept, additional optimization pressure is removed. In practice, we found that this threshold was only reached less than 1\% of the time during training.

\begin{figure*}[h!]
    \centering
    \includegraphics[width=\linewidth]{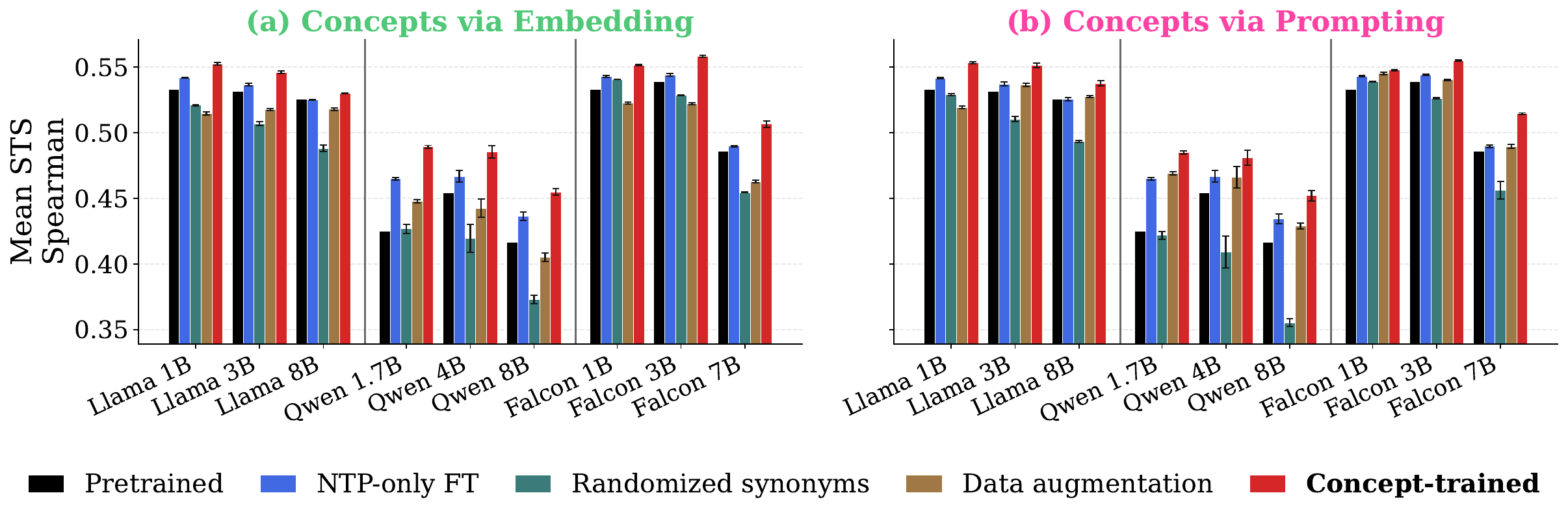}
    \caption {\textbf{Concept-trained models align more closely with human semantic intuition.} Results are for models trained on C4 and averaged over three training seeds, with error bars showing the standard deviation across seeds. Concept models are shown at $\lambda=1$. Randomized-synonym models are shown at $\lambda=0.25$ and data-augmentation models at 1 epoch, which are the best-performing settings for each baseline.}
  \label{fig:sim}
\end{figure*}

\section{Experiments \& Results}
\label{sec:exp}

We evaluate our models on four fronts: alignment with human semantic similarity judgments (Section \ref{sec:similarity}), general-purpose embedding tasks (Section \ref{sec:mteb}), downstream multiple-choice tasks (Section \ref{sec:downstream}), and NTP perplexity (Section~\ref{sec:ntp}). To align our evaluation with the context length used for post-training, we restrict the evaluation to short-context tasks and defer longer-context training and evaluation to future work (see Section~\ref{sec:limitations}).

Every model is post-trained with three random seeds, and all figures report the mean and standard deviation over those seeds. For the sake of visualization, the main text reports models trained on C4; the corresponding OpenWebText results are given in Appendix~\ref{appendix:openwebtext} and show the same trends.

\subsection{Increased Semantic Understanding}
\label{sec:similarity}

We evaluate alignment with human semantic judgments using the test splits of 9 Semantic Textual Similarity (STS) tasks from the Massive Text Embedding Benchmark (MTEB) ~\cite{enevoldsen2025mmtebmassivemultilingualtext, muennighoff2022mteb}. Specifically, we use the English subset of MTEB v2 which contains the STS tasks STS12, STS13, STS14, STS15, STS17, STS22.v2, STSBenchmark, SICK-R, and BIOSSES~\cite{10.5555/2387636.2387697, Agirre2013SEM2S, bandhakavi-etal-2014-generating, bicici-2015-rtm, cer-etal-2017-semeval, chen-etal-2022-semeval, huggingface:dataset:stsb_multi_mt, marelli-etal-2014-sick, 10.1093/bioinformatics/btx238}.

For each sequence, we obtain a representation by mean-pooling the final-layer hidden states and applying L2 normalization. MTEB's library then computes cosine similarity between sentence representations for each pair and reports Spearman correlation with the human similarity judgments. We then average the score reported using MTEB on the test splits of each of the STS datasets.

Figure~\ref{fig:sim} depicts the STS results, showing that \textbf{concept-trained models achieve higher Spearman correlation with human semantic similarity judgments than all baselines}. Spearman correlation also increases monotonically with concept weight $\lambda$, indicating that \textbf{stronger concept supervision yields representations that are increasingly aligned with human semantic judgments} (Appendix~\ref{appendix:sts_weight}). In contrast, the randomized-synonym baseline peaks at $\lambda=0.25$ and deteriorates as $\lambda$ increases, suggesting that while a small amount of noise may act as regularization, excessive noise ultimately harms semantic coherence (Appendix~\ref{appendix:sts_weight}). This contrast supports our claim that the gains come from semantically meaningful concept sets, not from adding noise.

Larger models generally achieve slightly lower Spearman correlation than their smaller counterparts, consistent with \citet{shani2026tokens}, who found that scaling does not correlate with human semantic judgments. One possible explanation is methodological: we extract embeddings by mean-pooling the final layer, whereas larger models may encode semantic information more strongly in intermediate layers. Nevertheless, we use final-layer mean pooling consistently across all models, following prior work on decoder embeddings~\cite{muennighoff2022sgpt, behnamghader2024llmvec, muennighoff2025generative}, to ensure comparability.

\subsection{Generalization Across Embedding Tasks}
\label{sec:mteb}

\begin{figure*}[h]
  \centering
  \includegraphics[width=\linewidth]{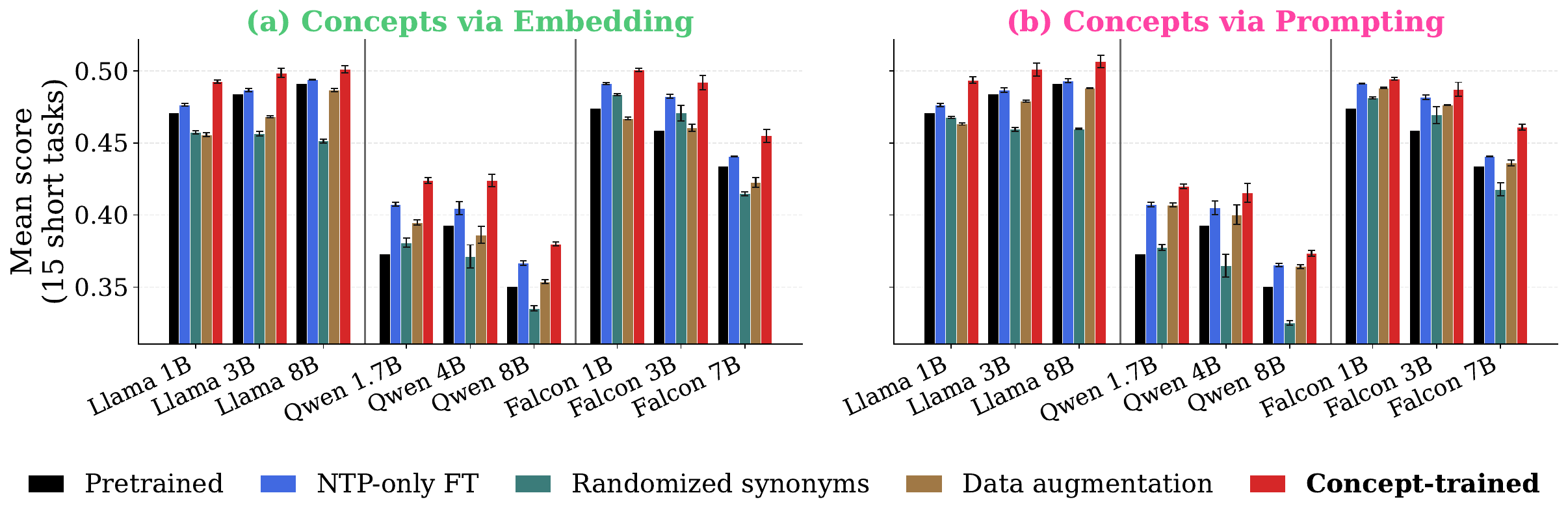}
  \caption{\textbf{Concept-trained models outperform all baselines when averaged across short-context MTEB tasks spanning classification, clustering, pair-classification, and reranking.} Scores are averaged over all 15 of these tasks, for models trained on C4 over three training seeds. Error bars show the standard deviation across seeds. Concept models are shown at $\lambda=1$, randomized-synonym models at $\lambda=0.25$, and data-augmentation models at 1 epoch, which are the best-performing settings for each baseline mirroring Figure~\ref{fig:sim}.}
  \label{fig:mteb}
\end{figure*}

We evaluate whether concept training improves the quality of learned \textit{embeddings} using all MTEB (eng, v2) tasks whose inputs fit within 256 tokens, matching our short-context training regime (Section~\ref{sec:data}). This results in 15 tasks spanning 7 classification, 3 pair-classification, 1 reranking, and 4 clustering benchmarks across domains, including product reviews, social media, technical Q\&A, and scientific abstracts. All evaluations operate directly on frozen last-layer mean-pooled embeddings through the MTEB library, which applies a trained linear probe for classification, cosine similarity for pair-classification and reranking, and unsupervised $k$-means for clustering.

We find that \textbf{concept training improves short-context representation over every baseline}, consistent across classification, pair-classification, reranking, and 3 of the 4 clustering tasks (Figure~\ref{fig:mteb}; per-category results in Appendix~\ref{appendix:mteb_by_task}). The sole exception is ArXivHierarchicalClusteringS2S~\cite{enevoldsen2025mmtebmassivemultilingualtext, muennighoff2022mteb}, where concept models and the best baseline fall within measurement noise of each other (mean difference -0.12 V-measure points across models and probes). Notably, this is the only hierarchical clustering benchmark in our suite, making it qualitatively distinct from the remaining tasks and potentially less aligned with the local semantic abstractions encouraged by concept training. Nevertheless, the overall pattern across the other 14 tasks remains strongly positive. As with STS, larger Qwen and Falcon models obtain lower absolute scores, likely for the same embedding-extraction reasons discussed in Section~\ref{sec:similarity}, though concept models still consistently outperform their corresponding baselines.

Taken together, these results indicate that \textbf{concept supervision yields better general-purpose representations}. The gains hold whether a task is solved by a trained probe, a cosine-similarity threshold, or unsupervised clustering, suggesting the improvement lies in the geometry of the embeddings themselves.

\subsection{Improved Downstream Reasoning}
\label{sec:downstream}

\begin{figure*}[h!]
    \centering
    \includegraphics[width=\linewidth]{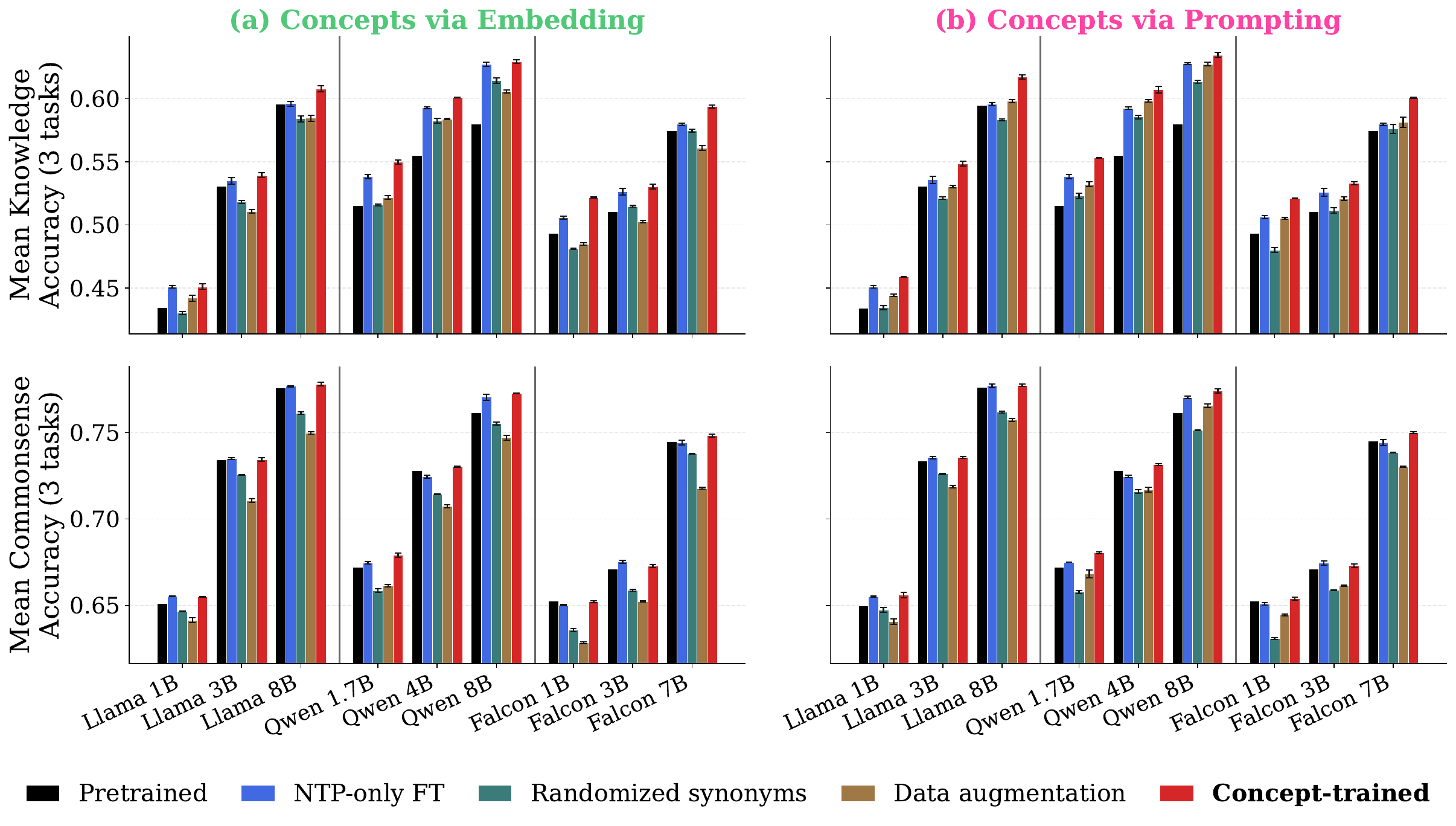}
    \caption{\textbf{Top}: Concept-trained models perform better on knowledge reasoning tasks (ARC-Challenge, ARC-Easy, OpenBookQA) compared to all baselines. \textbf{Bottom}: Concept-trained models match accuracy on commonsense tasks (HellaSwag, PIQA, WinoGrande) compared to the best-performing baseline models. Results are for models trained on C4 and averaged over three training seeds, with error bars showing the standard deviation across seeds. Concept models are shown at $\lambda=1$. Randomized-synonym models are shown at $\lambda=0.25$ and data-augmentation models at 1 epoch, which are the best-performing settings for each baseline.}
    \label{fig:reasoning_commonsense}
\end{figure*}

In line with our short-context training regime, we evaluate our models on six multiple-choice downstream benchmarks using the Language Model Evaluation Harness~\cite{eval-harness}: ARC-Challenge, ARC-Easy, OpenBookQA, HellaSwag, PIQA, and WinoGrande~\cite{clark2018thinksolvedquestionanswering, OpenBookQA2018, zellers2019hellaswag, Bisk2020, winogrande}. We group these benchmarks into two categories: \emph{knowledge reasoning} (ARC-Challenge, ARC-Easy, and OpenBookQA), which emphasize scientific and factual reasoning, and \emph{commonsense reasoning} (HellaSwag, PIQA, and WinoGrande), which probe physical, linguistic, and situational commonsense.

The top part of Figure~\ref{fig:reasoning_commonsense} shows length-normalized accuracy averaged across the \textit{knowledge reasoning} tasks. As expected, performance improves with model size across all methods. Moreover, \textbf{concept models consistently outperform all baselines on knowledge reasoning benchmarks}, supporting the hypothesis that the gains arise from the concept objective itself rather than from alternative factors controlled for in the baselines. Higher concept weights generally lead to better performance on these benchmarks; we include results comparing reasoning performance against concept weight in Appendix~\ref{appendix:downstream_weight}.

The bottom of Figure~\ref{fig:reasoning_commonsense} shows accuracy averaged across the commonsense reasoning tasks.\footnote{We report length-normalized accuracy where available; for WinoGrande, this is equivalent to raw accuracy.} Here we find no benefit from concept supervision: \textbf{concept models perform on par with the pure NTP baselines}. We hypothesize that this asymmetry in knowledge versus commonsense reasoning reflects a difference in what the two task categories demand. Knowledge reasoning relies on explicit factual and lexical content, where grouping content words with their semantic neighbors plausibly sharpens the representations a model draws on. Commonsense reasoning instead relies on implicit physical and situational expectations that are not well captured by content-word synonymy, leaving little for the concept objective to exploit.



\subsection{NTP Perplexity Analysis}
\label{sec:ntp}

\begin{figure*}[h!]
  \centering
  \includegraphics[width=\linewidth]{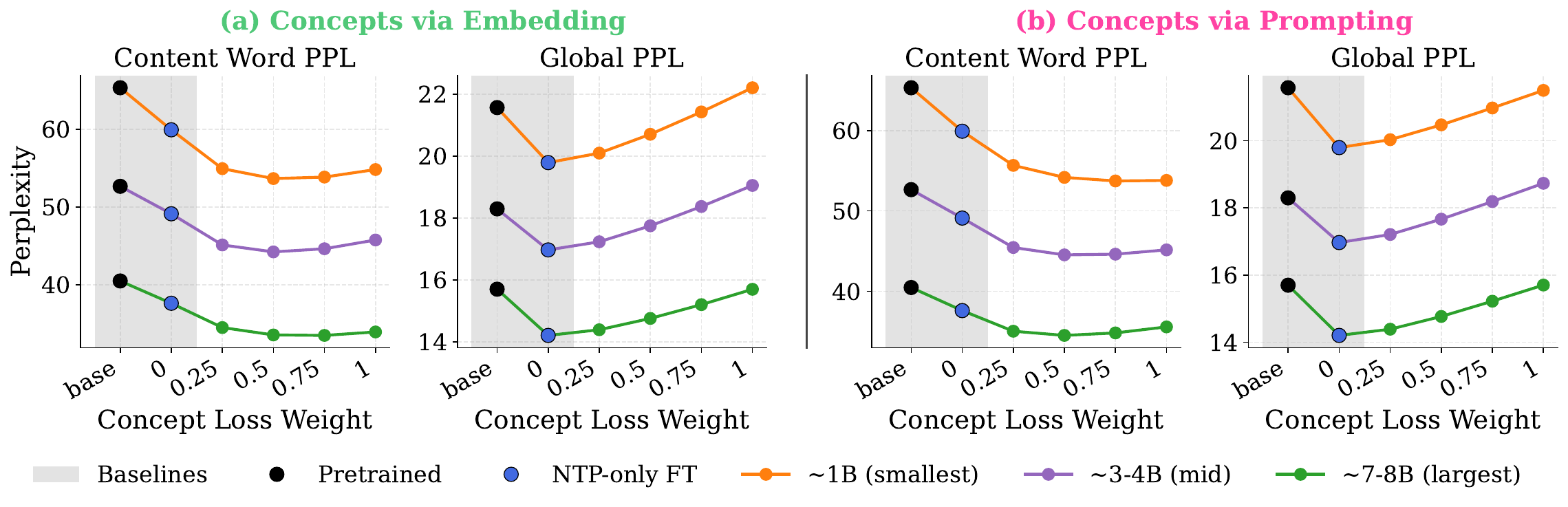}
  \caption {Concept-trained models achieve substantially lower content word perplexity than both the pretrained and NTP-only baselines, while global perplexity degrades compared to the NTP-only baseline but is on par with the pretrained model. Results are for models trained on C4, averaged over three training seeds and across the three model families, grouped by model size. We omit error bars because absolute perplexity values differ substantially across families due to differences in their base models; per-family plots are reported in Appendix~\ref{appendix:ppl_by_family}.}
  \label{fig:ntp}
\end{figure*}

\begin{figure*}[h!]
  \centering
  \includegraphics[width=\linewidth]{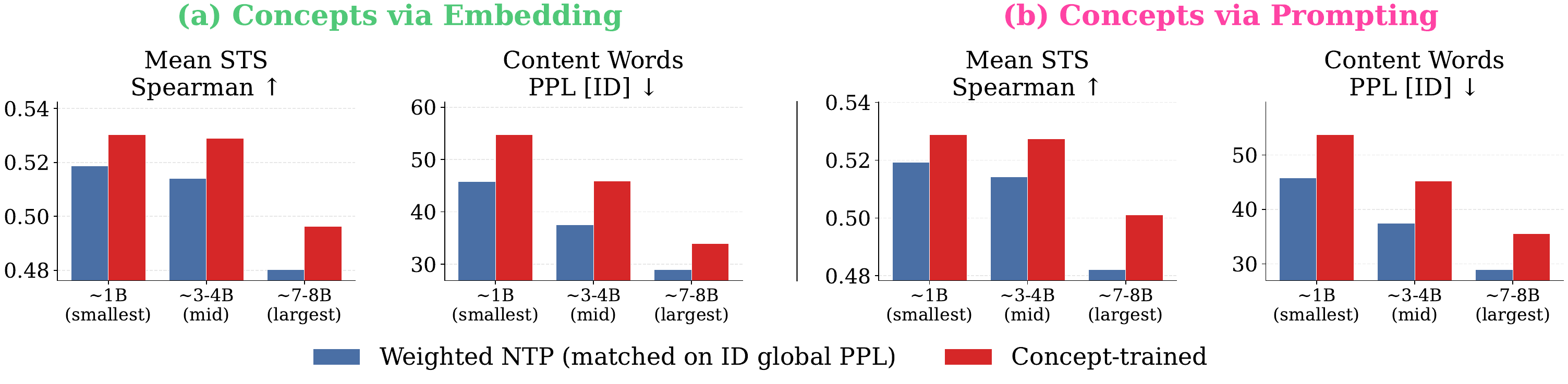}
  \caption {Concept-trained models outperform weighted-NTP models on STS Spearman correlation while weighted-NTP models achieve lower content word perplexity. Each weighted-NTP model is selected to match its corresponding concept model on in-domain global perplexity, isolating the effect of the training objective. Results are for models trained on C4, averaged over model families (see individual plots in Appendix~\ref{appendix:wce_by_model}).}
  \label{fig:wce}
\end{figure*}

\begin{figure*}[h!]
  \centering
  \includegraphics[width=\linewidth]{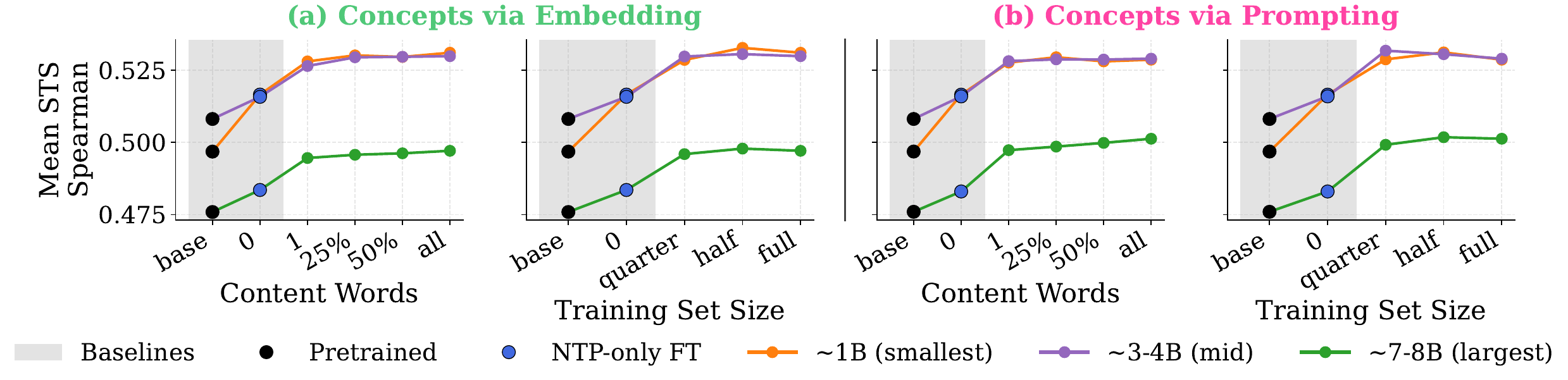}
  \caption {Concept training saturates at small data amounts on STS along both axes we vary. For the training set size plots, ``0'' denotes training on the full dataset at $\lambda=0$. All models other than the pretrained and pure-NTP baselines use $\lambda=1$. In these plots, we use models trained on C4 and aggregate over three training seeds and model families (see individual plots in Appendix~\ref{appendix:saturation_by_family}).}
  \label{fig:max_data_ablation}
\end{figure*}

We analyze NTP perplexity on both in-domain (ID) held-out samples and out-of-distribution (OOD) samples from the alternate corpus (C4 or OpenWebText). Alongside standard perplexity, we report \textit{content word perplexity}, computed only over content-word tokens. This metric isolates performance on semantically meaningful items (approximated as complete-word single-token nouns, verbs, and adjectives), excluding the function words and punctuation that are typically easier to predict:
\[
\text{Content Word PPL} =
\exp\left(
\frac{1}{|\mathcal{T}|}
\sum_{t \in \mathcal{T}} \text{NLL}(t)
\right)
\]
\noindent where $\mathcal{T}$ is the set of content-word tokens and $\mathrm{NLL}(t)$ is the negative log-likelihood of token $t$.

Figure~\ref{fig:ntp} shows results aggregated over models and datasets for ID data; OOD results follow the same trends (Appendix~\ref{appendix:ntp_ood}). \textbf{Content word perplexity decreases monotonically with concept weight across both ID and OOD data and both concept extraction methods.} Thus, concept supervision improves not only downstream behavior but also NTP at content word positions, despite training to spread probability across many plausible continuations. Additionally, content word perplexity is substantially higher than global perplexity across all conditions, indicating that content words are the harder tokens to predict in each sequence. The improvements from concept supervision are concentrated on these semantically loaded positions, where our models outperform both the pretrained models and NTP post-trained baselines.

For global perplexity, concept models perform comparably to the pretrained baseline but trail the NTP baseline by $\approx$1-2 points. We view this as a tradeoff: \textbf{concept supervision shifts capacity toward semantically informative tokens at a modest cost on function words and other positions that contribute less to sentence meaning.}

Appendix~\ref{appendix:ppl_by_family} depicts per-family plots, which show the same trends for content-word and global ID perplexity.

\section{Ablations}
\label{sec:ablations}

We perform two ablations, testing whether concept supervision drives the gains or the increased weight assigned to content words during training (Section~\ref{sec:weight_ntp}), and studying how the amount of concept supervision along different axes affects STS Spearman correlation (Section~\ref{sec:small_data}).

\subsection{Increased Concept-Word Weight via NTP}
\label{sec:weight_ntp}

To test whether gains come from the concept objective or simply emphasizing content words, we train pure-NTP models that have a higher loss weight on content words, sweeping weights over $\{2, 5, 10, 20, 50, 100\}$ while keeping other tokens at weight 1.

For an unbiased comparison controlling for language modeling quality, Figure~\ref{fig:wce} matches each concept model ($\lambda=1$) with the weighted-NTP model closest to it in global ID perplexity, controlling for general predictive quality. This yields a consistent choice of content-word weight 5 across all families and sizes.

Under this matched comparison, weighted-NTP outperforms our concept-trained models on content-word perplexity but performs worse on STS correlation. This is expected: it directly optimizes each content token, while the concept objective distributes probability over synonyms and cannot match NTP by design. However, this sharper probability does not improve semantic representations, as concept models align better with human similarity judgments. This indicates that \textbf{gains come from the concept objective, not merely emphasizing content word positions}.

\subsection{Saturation at Small Data Amounts}
\label{sec:small_data}

To assess the role of concept supervision, we vary it along two axes: the fraction of supervised content words per sequence (all, 50\%, 25\%, a single word), and the training set size (full, 50\%, 25\%).

\textbf{Reducing the supervised content words has little effect on STS performance.} Applying concept loss to a single content word per sequence yields Spearman correlation comparable to applying it to all content words (Figure~\ref{fig:max_data_ablation}). The concept signal thus saturates quickly: \textbf{one concept per sequence suffices to convey the inductive bias}, and supervising many synonym sets may introduce overfitting.

\textbf{Reducing training set size also has little effect.} Spearman correlation stays essentially flat as we cut the training set to 50\% and 25\%, indicating that concept post-training takes effect with only a small amount of data. This finding is consistent with prior work on the data-efficiency of supervised fine-tuning~\cite{zhou2023lima, muennighoff-etal-2025-s1, chen2024alpagasus}.

Similar saturation patterns hold for content word and global perplexity (Appendix~\ref{appendix:saturation}) and for individual model families (Appendix~\ref{appendix:saturation_by_family}).

\section{Conclusions and Future Directions}

We introduce a self-supervised, concept-based training objective that complements standard next-token prediction. Across three model families and three sizes, concept-trained models align more closely with human semantic similarity judgments, improve on short-context classification, clustering, and reranking tasks, strengthen knowledge reasoning, and reduce perplexity on semantically informative tokens. Comparisons against pretrained, pure-NTP, randomized-synonym, and data-augmented NTP baselines indicate that these gains stem from the concept objective itself rather than confounding factors. Notably, a small increase in global perplexity buys substantial gains on semantically meaningful tokens and downstream performance, suggesting that semantically targeted metrics can be more informative than aggregate perplexity alone.

Future work could incorporate concept supervision during pretraining, extend it to longer contexts, or apply it to multi-token expressions and multilingual data. Future training could also utilize qualitatively different or larger-scale corpora. More broadly, we view concept-level objectives as a step toward LLMs whose representations more faithfully reflect the conceptual organization underlying human language.

\section*{Limitations}
\label{sec:limitations}

Although we show promising results with concept training, our study has several limitations. First, our concept supervision is applied only to nouns, verbs, and adjectives that are a single token and a complete word. This restricts our training paradigm to a narrow prediction setting and may not fully capture the benefits of concept-level learning across longer and more diverse spans of text. Extending supervision to different types of tokens within sequences could yield stronger effects.

Second, our methods for constructing concept sets rely on either the decoder model's embedding structures or LLM-generated synonym clusters. While scalable, these approaches may not perfectly reflect human semantic organization, and alternative sources, such as human annotations, could provide more reliable or linguistically grounded concept groupings. 

Third, our experiments focus exclusively on decoder-based models with no instruction finetuning, which limits our ability to evaluate performance under instruction prompting for downstream tasks. As a result, our evaluation tasks in this work may not capture broader ways in which concept training could be applied to improve language modeling ability.

Additionally, because we train on data truncated to 256 tokens, our models are not well-suited for long-context tasks such as retrieval. Future work can explore concept training for longer contexts.

We also acknowledge that our models do not achieve state-of-the-art embedding performance in Figures~\ref{fig:sim} and \ref{fig:mteb}. This is expected given that we are using relatively small decoder models without task-specific finetuning. Our goal in this work is to show that concept training improves semantic understanding \textit{in the absence of} task-specific embedding finetuning. When we apply contrastive finetuning on top of a concept post-trained model, the effect of contrastive finetuning dominates our prior post-training. We include details of our contrastive finetuning experiments in Appendix~\ref{appendix:contrastive}, and we leave to future work to investigate more sophisticated methods of composing concept training with contrastive finetuning.

\section*{Ethical Considerations}

Training models with our concept loss formulation could potentially lead to overgeneralizing and overextending concept boundaries. The approach we describe in this proposal involves LLMs creating concept sets without human input, which can be dangerous in amplifying biases and stereotypes present in training data. We advise users to be careful of potentially harmful associations generated in the concept dataset creation.

As with NTP models, concept-trained models may behave in an unwanted or harmful manner, such as producing hallucinations. In this work, we did not explore using our concept-trained models for unconditioned generation, and thus the severity of these risks is unknown for our models. We caution users to always check model outputs.

Disclosure: LLMs were used to assist with coding and refine the text of this paper.


\bibliography{may_custom.bib}

\clearpage
\appendix

\section{Prompt to Extract Concept Sets}
\label{appendix:instruct_prompt}

This is the prompt fed to the instruction model to select the subset of contextual synonyms from the top 100 decoded tokens for each decoder model. We use deterministic sampling for reproducibility and a repetition penalty of 1.1.

\begin{quote}
\small
\ttfamily
Find contextual synonyms for the word \{target\} in this text: \{input\_sequence\}

Available tokens: \{decoded\_tokens\}

Instructions:

- Find ALL possible synonyms from the available tokens that could replace \{target\} in this context

- Return ONLY a comma-separated list of synonyms from the available tokens, surrounded by square brackets

- Include every relevant synonym

- NO duplicates allowed

- NO explanations or extra text

- If no synonyms found, return: []

Example format: [word1, word2, word3]

Synonyms for \{target\}:
\end{quote}

\section{Concept Set Statistics}
\label{appendix:concept_stats}

This section reports the size of the concept sets produced by each stage of the pipeline described in Section~\ref{sec:data}. All statistics are computed over the full 10,000 sequences per corpus, before the 80/10/10 split.

Table~\ref{tab:content_words} reports the number of content words per sequence. Because this stage depends only on the tokenizer and the spaCy part-of-speech tags, it is identical for all three model sizes within a family, so we report one row per family. Sequences are truncated to 256 tokens and average 190 tokens on C4 and 251 tokens on OpenWebText, so content words make up roughly 24\% of tokens per sequence in both corpora. The much larger standard deviation on C4 reflects that corpus's shorter and more variable documents.

Table~\ref{tab:concept_sizes} reports, for each model and corpus, the size of the candidate pool and the size of the final concept set under each generation method. The candidate pool is the top 100 next-token predictions after discarding candidates that are not alphanumeric whole words or that repeat the original token. The pool averages 91--95 candidates per content word and is shared by both the embedding and prompting methods.

Two aspects of the final concept sets are worth noting. First, the two methods are bounded differently: the embedding method is explicitly capped at 10 synonyms per content word, whereas the prompting method is limited by the length of the instruction model's response to 20 tokens. As a result, the concept sets for the prompting method are occasionally slightly larger than 10 but are in practice smaller on average. Second, the prompting method is markedly less reliable at small scale. Falcon 3 1B returns an empty list for roughly two thirds of content words (31--35\% coverage, versus 74--95\% for every other model), which is why its mean concept set size is far below that of the other models. The embedding method, which does not depend on instruction-following ability, is stable across all nine models.

\section{Results on OpenWebText}
\label{appendix:openwebtext}

\begin{figure*}[h!]
  \centering
  \includegraphics[width=\linewidth]{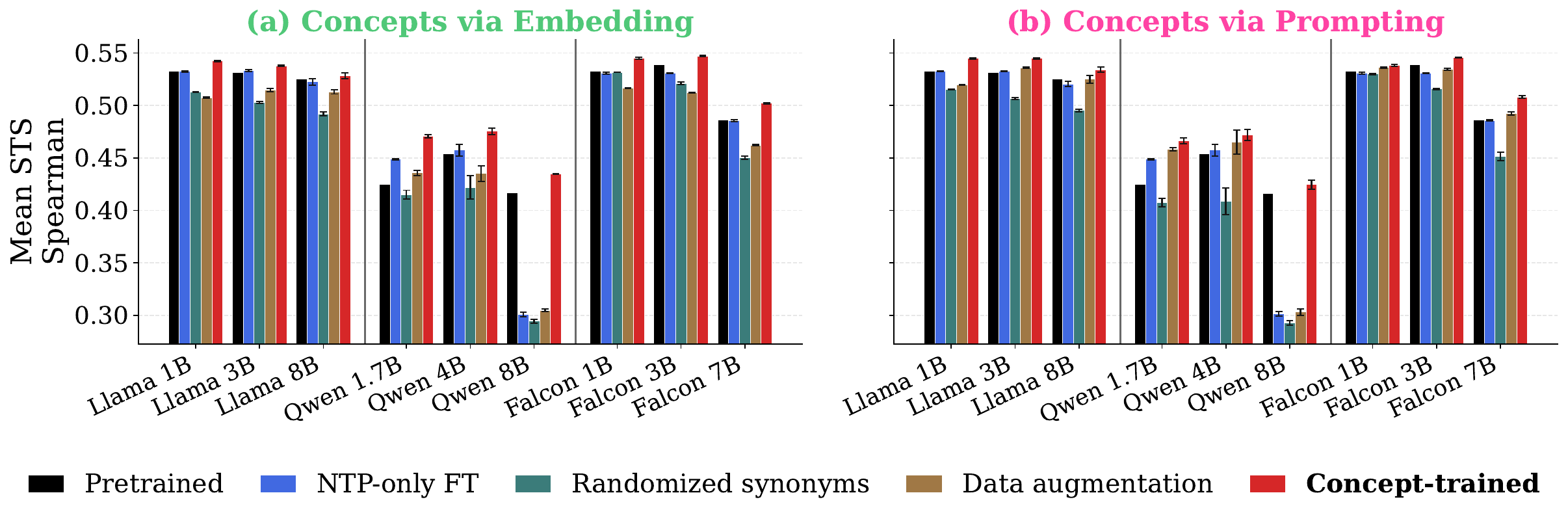}
  \caption {OpenWebText counterpart of Figure~\ref{fig:sim}. Concept-trained models align more closely with human semantic intuition. Concept models are shown at $\lambda=1$, randomized-synonym models at $\lambda=0.25$, and data-augmentation models at 1 epoch.}
  \label{fig:sim_owt}
\end{figure*}

\begin{figure*}[h!]
  \centering
  \includegraphics[width=\linewidth]{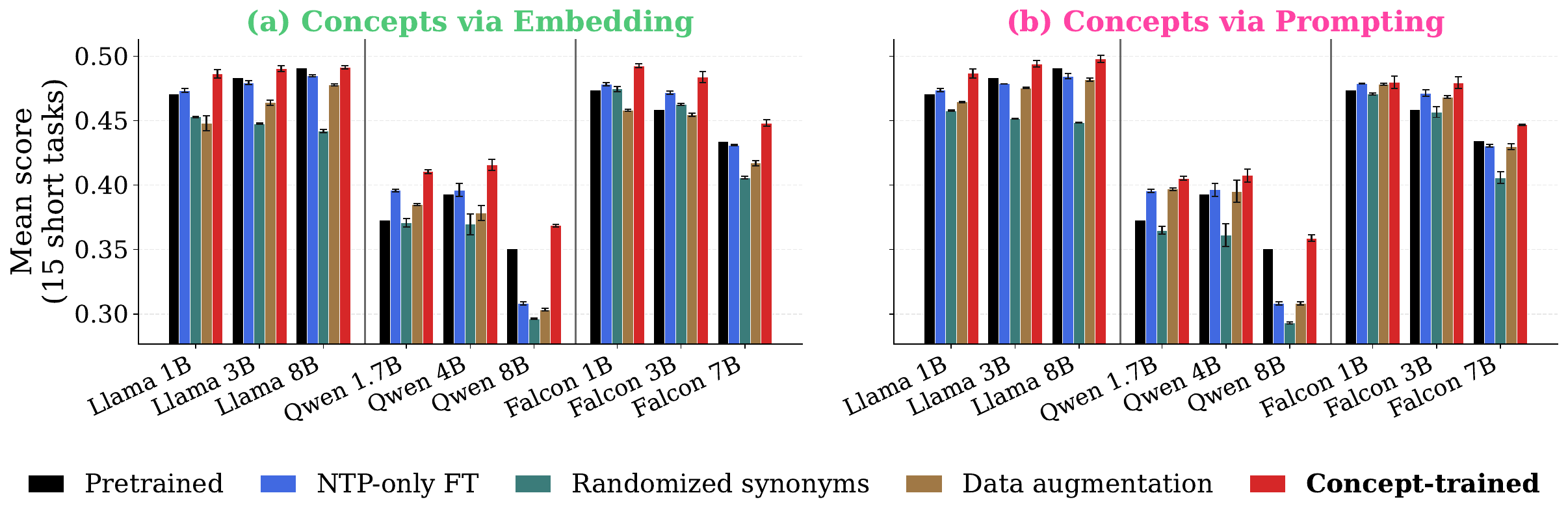}
  \caption {OpenWebText counterpart of Figure~\ref{fig:mteb}. Concept-trained models outperform all baselines when averaged across the 15 short-context MTEB tasks. Concept models are shown at $\lambda=1$, randomized-synonym models at $\lambda=0.25$, and data-augmentation models at 1 epoch.}
  \label{fig:mteb_owt}
\end{figure*}

\begin{figure*}[h!]
  \centering
  \includegraphics[width=\linewidth]{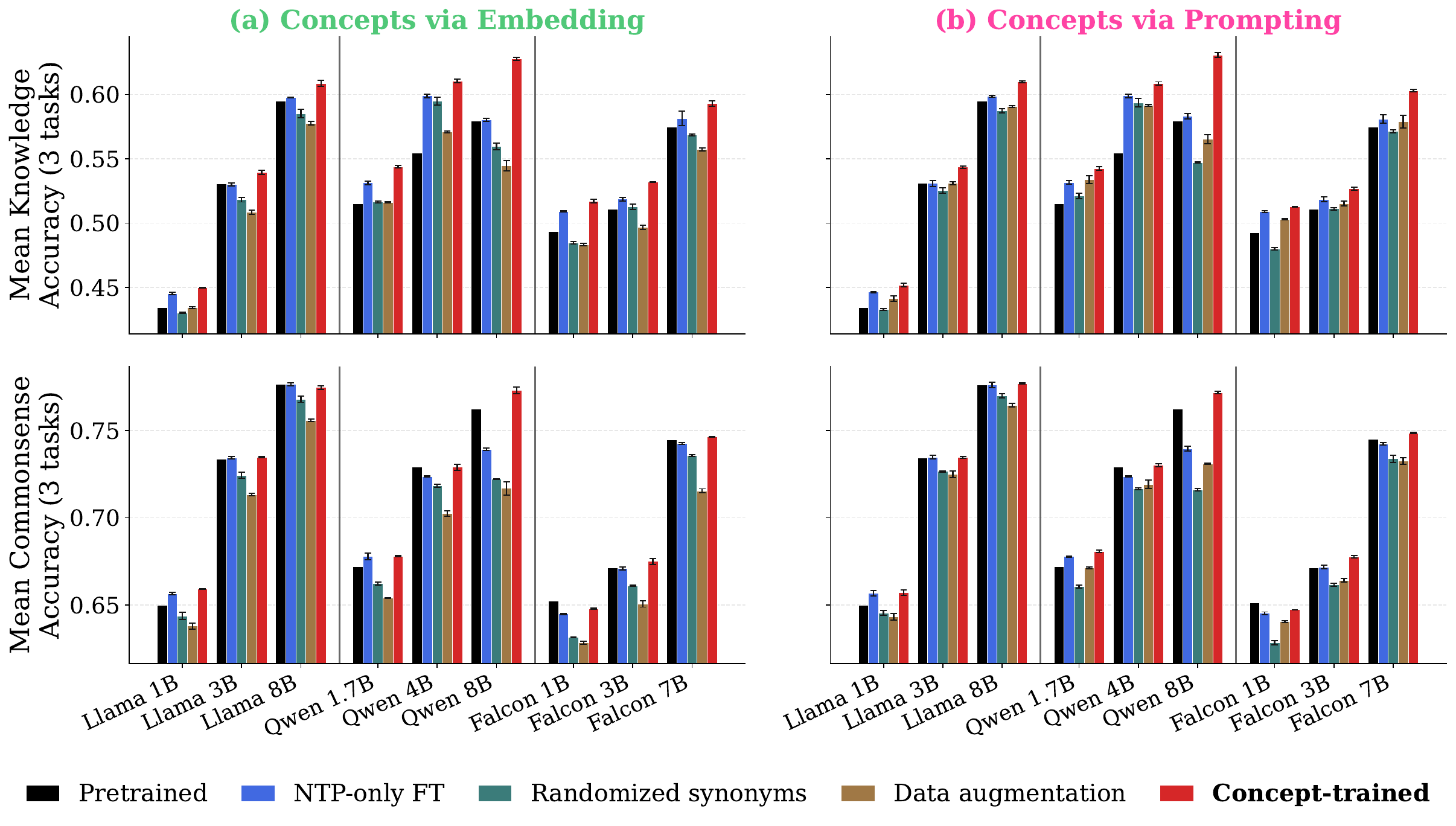}
  \caption {OpenWebText counterpart of Figure~\ref{fig:reasoning_commonsense}. \textbf{Top}: knowledge reasoning tasks (ARC-Challenge, ARC-Easy, OpenBookQA). \textbf{Bottom}: commonsense tasks (HellaSwag, PIQA, WinoGrande). Concept models are shown at $\lambda=1$, randomized-synonym models at $\lambda=0.25$, and data-augmentation models at 1 epoch.}
  \label{fig:reasoning_commonsense_owt}
\end{figure*}

\begin{figure*}[h!]
  \centering
  \includegraphics[width=\linewidth]{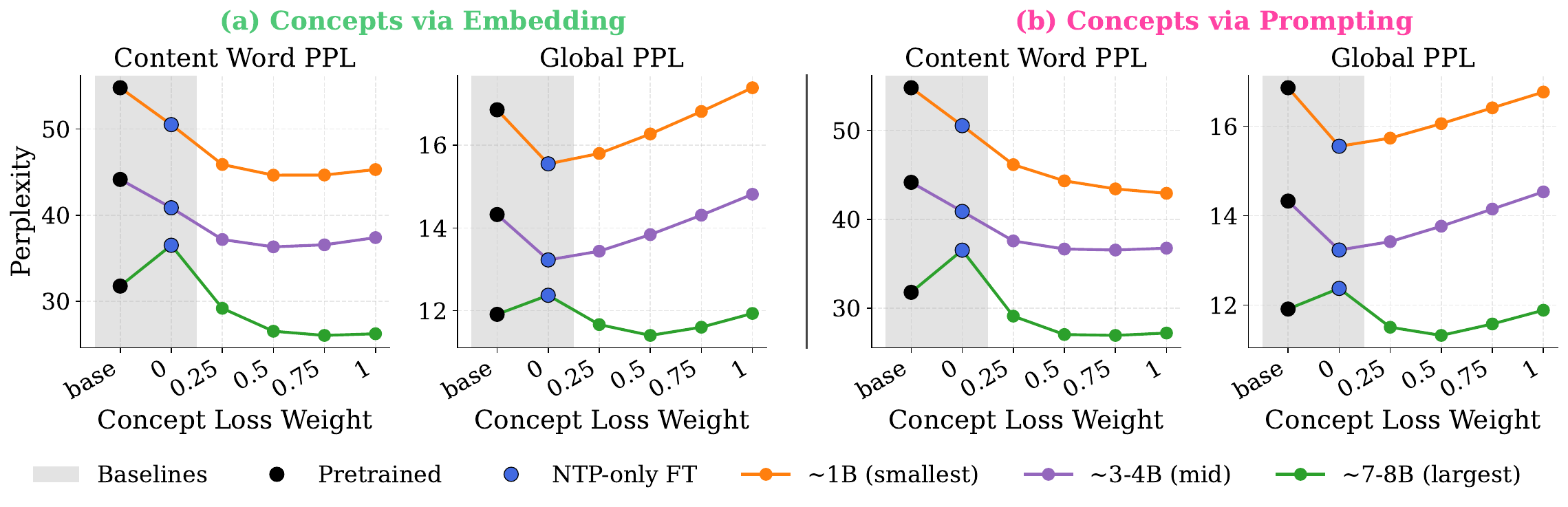}
  \caption {OpenWebText counterpart of Figure~\ref{fig:ntp}. Concept-trained models achieve substantially lower content word perplexity than both the pretrained and NTP-only baselines, while global perplexity degrades compared to the NTP-only baseline but performs on par with the pretrained model. Results are averaged across the three model families, grouped by model size; we omit error bars because absolute perplexity values differ substantially across families.}
  \label{fig:ntp_owt}
\end{figure*}

\begin{figure*}[h!]
  \centering
  \includegraphics[width=\linewidth]{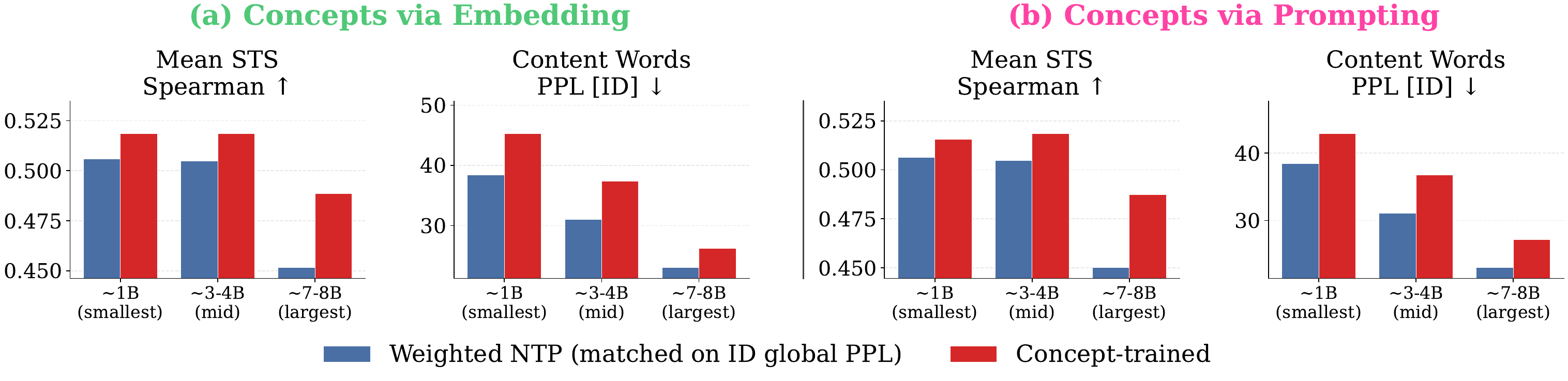}
  \caption {OpenWebText counterpart of Figure~\ref{fig:wce}. Concept-trained models outperform weighted-NTP models on STS Spearman correlation while weighted-NTP models achieve lower content word perplexity. Each weighted-NTP model is selected to match its corresponding concept model on in-domain global perplexity.}
  \label{fig:wce_owt}
\end{figure*}

\begin{figure*}[h!]
  \centering
  \includegraphics[width=\linewidth]{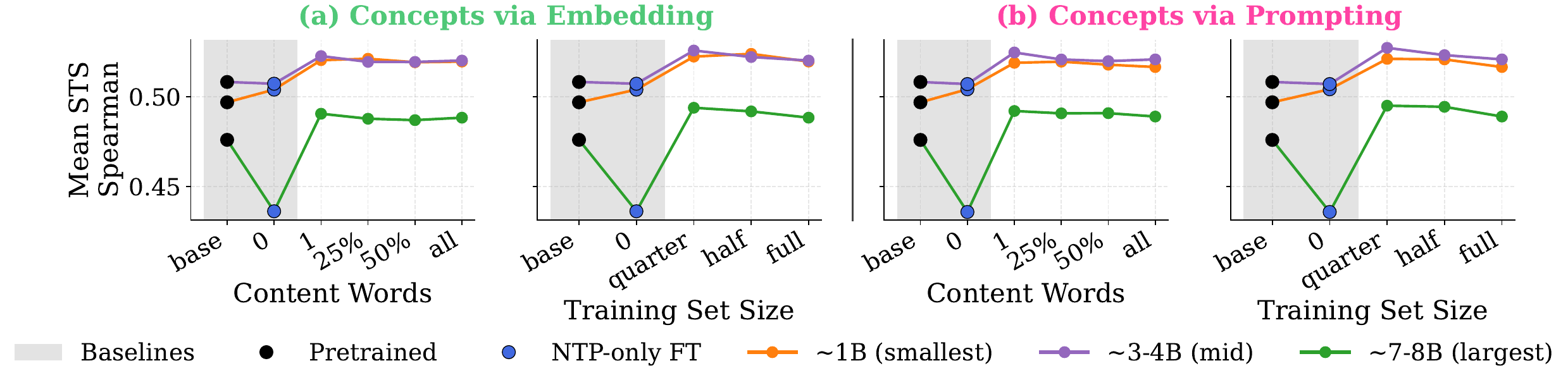}
  \caption {OpenWebText counterpart of Figure~\ref{fig:max_data_ablation}. Concept training saturates at small data amounts on STS along both axes we vary. For the training set size plots, ``0'' denotes training on the full dataset at $\lambda=0$. All models other than the pretrained and pure-NTP baselines use $\lambda=1$.}
  \label{fig:max_data_ablation_owt}
\end{figure*}

Figures~\ref{fig:sim_owt},~\ref{fig:mteb_owt},~\ref{fig:reasoning_commonsense_owt},~\ref{fig:ntp_owt},~\ref{fig:wce_owt}, and~\ref{fig:max_data_ablation_owt} repeat every main-text result for models post-trained on OpenWebText instead of C4. As in the main text, values are averaged over three training seeds, and error bars are the standard deviation across those seeds. All of the trends reported in the main text hold on this corpus as well.

\section{STS by Concept Weight}
\label{appendix:sts_weight}
\begin{figure*}[h!]
  \centering
  \includegraphics[width=\linewidth]{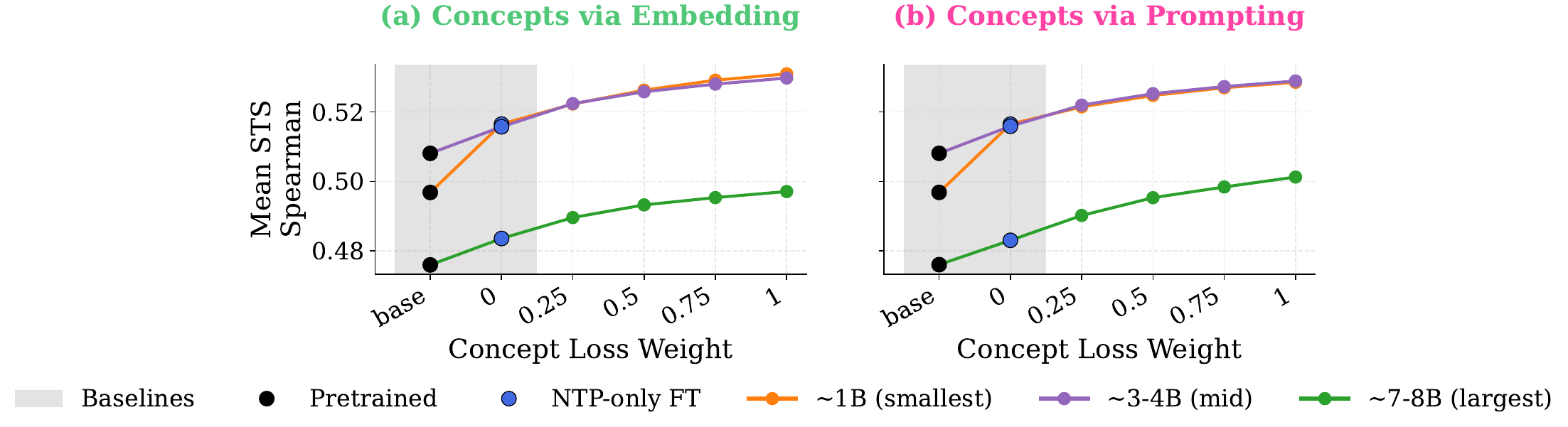}
  \caption {STS Spearman correlation increases as concept weight increases, indicating that a higher degree of concept supervision allows models to align more strongly with human semantic intuition.}
  \label{fig:sts_line}
\end{figure*}

We show results for STS Spearman correlation separated by concept weight, aggregated over model families in Figure~\ref{fig:sts_line}.

\begin{figure*}[h!]
  \centering
  \includegraphics[width=\linewidth]{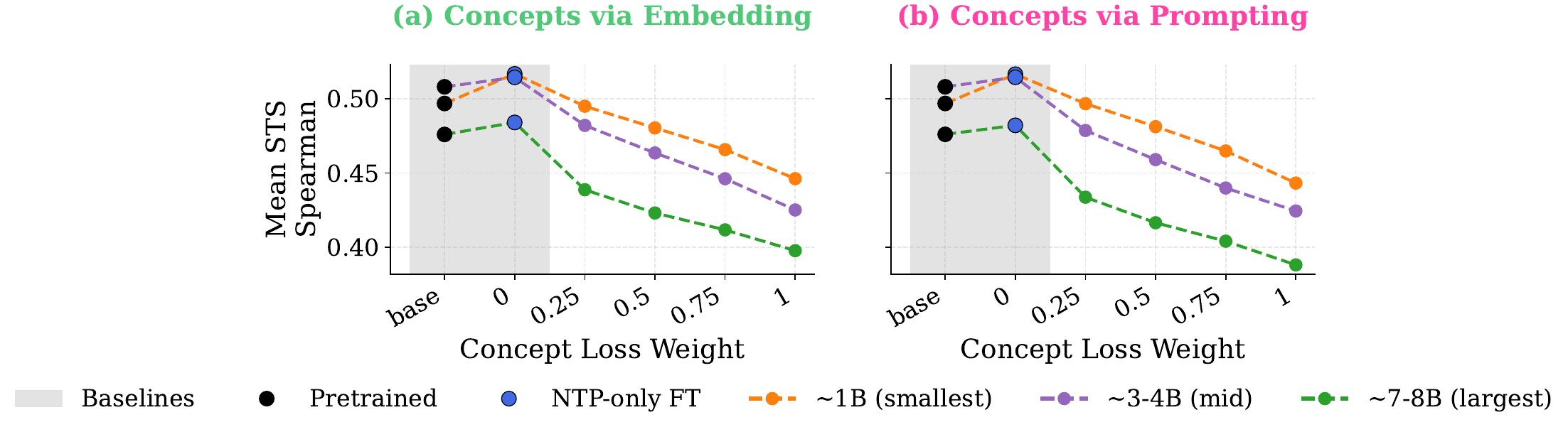}
  \caption {STS Spearman correlation decreases as concept weight increases for the models with randomized synonym sets. Thus, without semantic groupings of concept sets, model representations become more damaged as we increase the weight on these noisy concept sets.}
  \label{fig:sts_line_rand}
\end{figure*}

We also show STS Spearman correlation results for the randomized synonym baseline models in Figure~\ref{fig:sts_line_rand}.

\section{Short-Context MTEB by Task}
\label{appendix:mteb_by_task}

\begin{figure*}[h!]
  \centering
  \includegraphics[width=\linewidth]{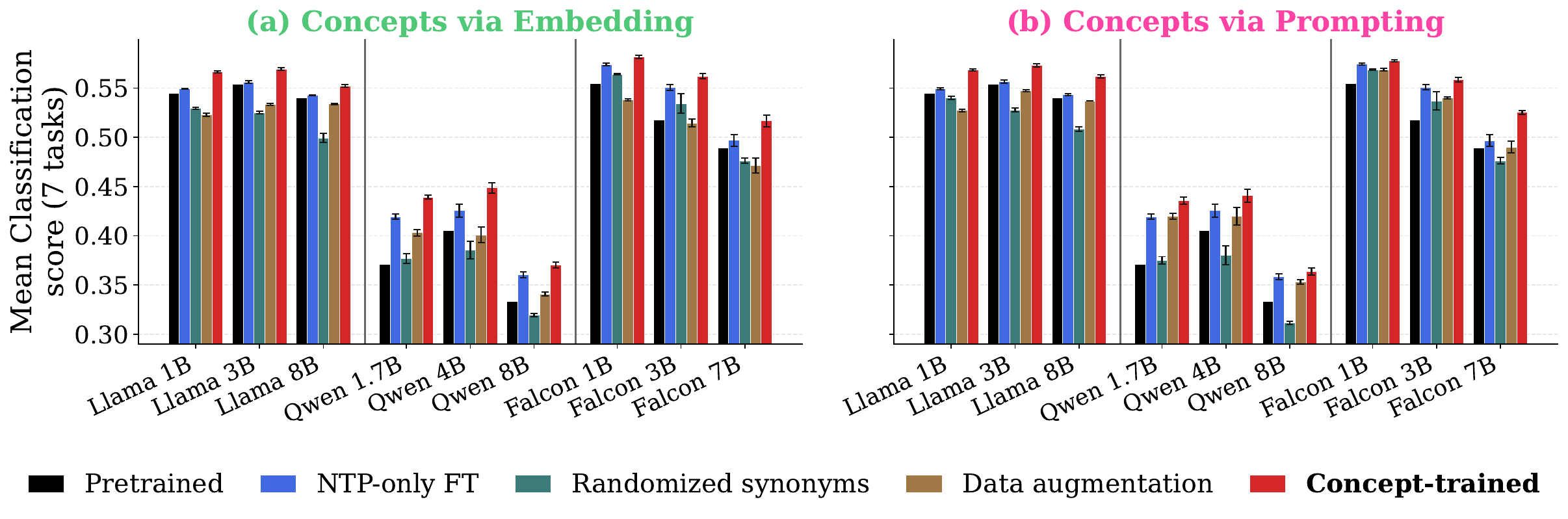}
  \caption {Concept-trained models outperform all baselines on classification tasks. Scores are averaged over 7 classification tasks for models trained on C4 over three training seeds; error bars show the standard deviation across seeds. Concept models are shown at $\lambda=1$, randomized-synonym models at $\lambda=0.25$, and data-augmentation models at 1 epoch.}
  \label{fig:classification}
\end{figure*}

\begin{figure*}[h!]
  \centering
  \includegraphics[width=\linewidth]{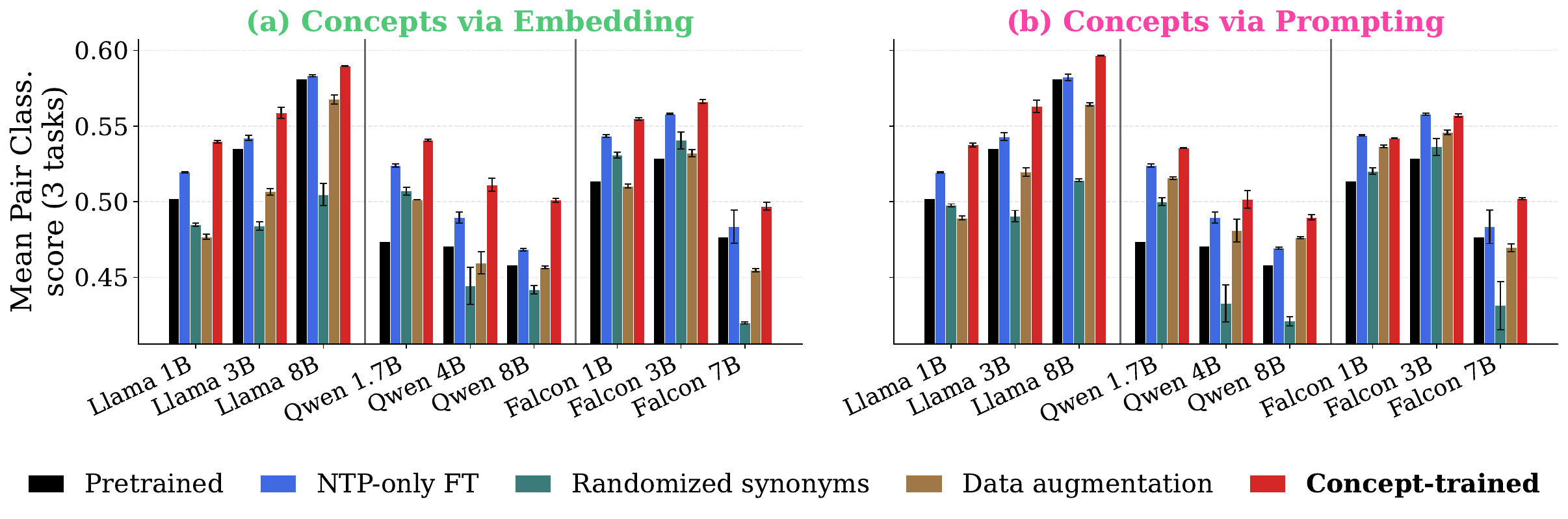}
  \caption {Concept-trained models outperform all baselines on pair-classification tasks. Scores are averaged over 3 pair-classification tasks for models trained on C4 over three training seeds; error bars show the standard deviation across seeds. Concept models are shown at $\lambda=1$, randomized-synonym models at $\lambda=0.25$, and data-augmentation models at 1 epoch.}
  \label{fig:pair_classification}
\end{figure*}

\begin{figure*}[h!]
  \centering
  \includegraphics[width=\linewidth]{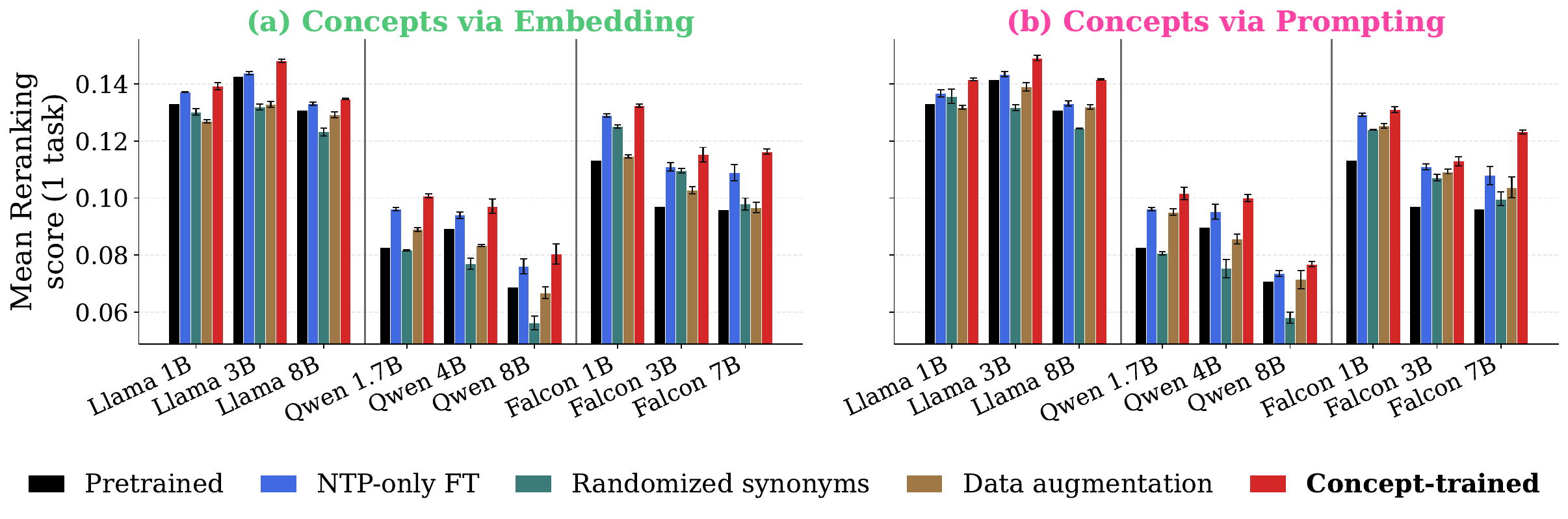}
  \caption {Concept-trained models outperform all baselines on the reranking task. Scores are for models trained on C4, averaged over three training seeds; error bars show the standard deviation across seeds. Concept models are shown at $\lambda=1$, randomized-synonym models at $\lambda=0.25$, and data-augmentation models at 1 epoch.}
  \label{fig:reranking}
\end{figure*}

\begin{figure*}[h!]
  \centering
  \includegraphics[width=\linewidth]{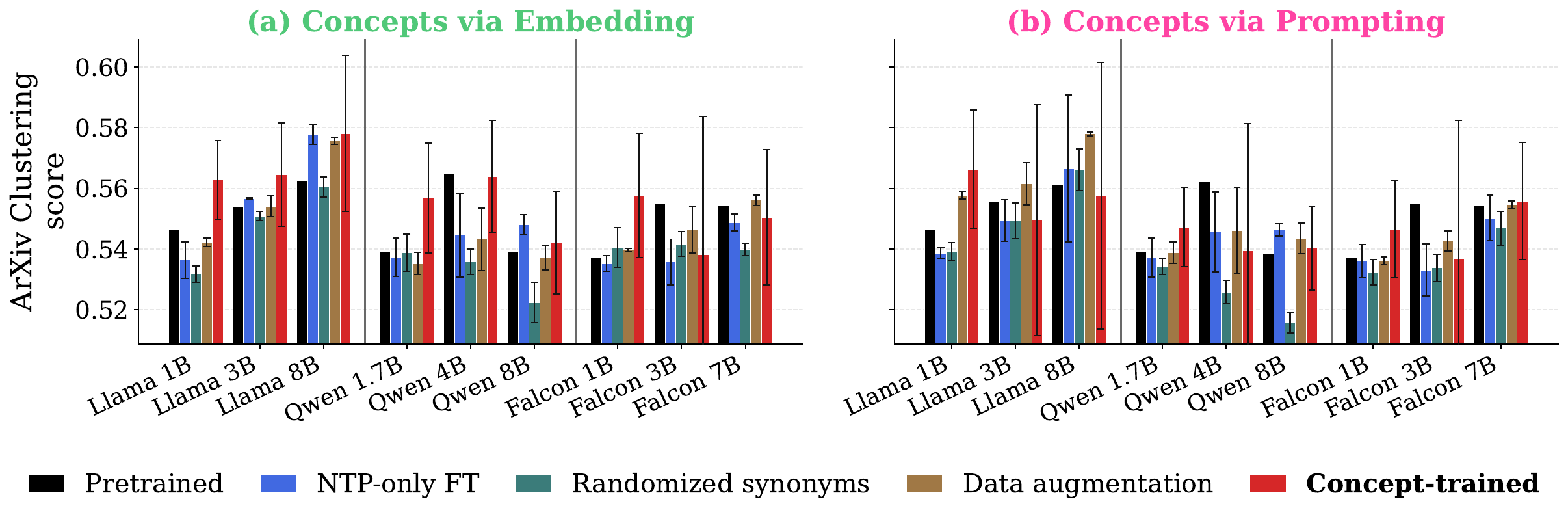}
  \caption {The arXiv clustering task cannot distinguish between the models. Scores are for models trained on C4, averaged over three training seeds; error bars show the standard deviation across seeds. ArXivHierarchicalClusteringS2S is the only hierarchical clustering benchmark in our suite, and MTEB scores it by bootstrapping V-measure across two label-hierarchy levels. This produces a within-run standard deviation of 3.3 points, which is far higher than the 0.6--0.9 of our other three clustering tasks. The spread shown here is therefore dominated by evaluation variance rather than by differences between training seeds, so we do not draw conclusions on model clustering performance given these method differences on this task. Concept models are shown at $\lambda=1$, randomized-synonym models at $\lambda=0.25$, and data-augmentation models at 1 epoch.}
  \label{fig:clustering_arxiv}
\end{figure*}

\begin{figure*}[h!]
  \centering
  \includegraphics[width=\linewidth]{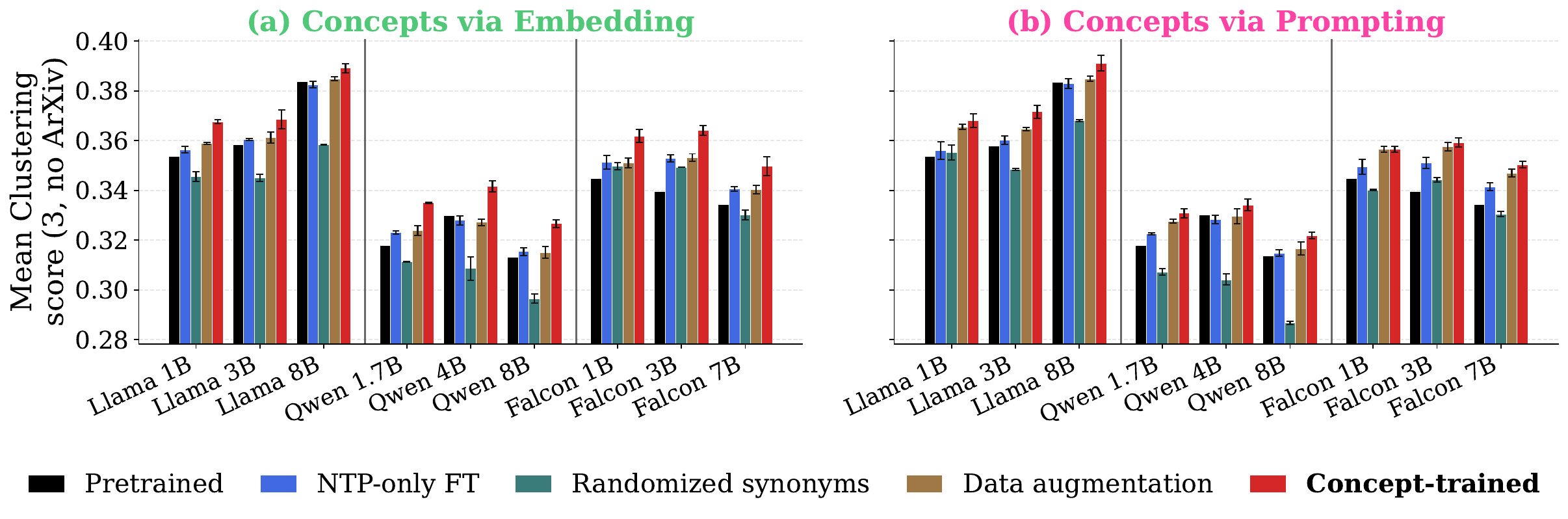}
  \caption {Concept-trained models generally outperform baselines on the remaining 3 clustering tasks, with small exceptions for Llama 1B, Falcon 1B, and Falcon 3B on the prompting side. Scores are averaged over these 3 tasks for models trained on C4 over three training seeds; error bars show the standard deviation across seeds. Concept models are shown at $\lambda=1$, randomized-synonym models at $\lambda=0.25$, and data-augmentation models at 1 epoch.}
  \label{fig:clustering_no_arxiv}
\end{figure*}

We report results for the 15 short-context MTEB~\cite{enevoldsen2025mmtebmassivemultilingualtext, muennighoff2022mteb} tasks, split by category:

\begin{itemize}
    \item \textbf{Classification (7 tasks; Figure~\ref{fig:classification}):}
    AmazonCounterfactualClassification~\cite{oneill-etal-2021-wish},
    Banking77Classification~\cite{casanueva-etal-2020-efficient},
    MassiveIntentClassification~\cite{FitzGerald2023},
    MassiveScenarioClassification~\cite{FitzGerald2023},
    MTOPDomainClassification~\cite{li-etal-2021-mtop},
    ToxicConversationsClassification~\cite{jigsaw-unintended-bias-in-toxicity-classification},
    TweetSentimentExtractionClassification~\cite{tweet-sentiment-extraction}.
    \item \textbf{Pair-classification (3 tasks; Figure~\ref{fig:pair_classification}):}
    SprintDuplicateQuestions~\cite{shah-etal-2018-adversarial},
    TwitterSemEval2015~\cite{xu-etal-2015-semeval},
    TwitterURLCorpus~\cite{lan-etal-2017-continuously}.
    \item \textbf{Reranking (1 task; Figure~\ref{fig:reranking}):}
    AskUbuntuDupQuestions~\cite{wang-etal-2021-tsdae-using}.
    \item \textbf{Clustering (4 tasks; Figure~\ref{fig:clustering_arxiv} for ArXivHierarchicalClusteringS2S and Figure~\ref{fig:clustering_no_arxiv} without):}
    ArXivHierarchicalClusteringS2S~\cite{enevoldsen2025mmtebmassivemultilingualtext, muennighoff2022mteb},
    MedrxivClusteringS2S.v2~\cite{enevoldsen2025mmtebmassivemultilingualtext, muennighoff2022mteb},
    StackExchangeClustering.v2~\cite{geigle:2021:arxiv},
    TwentyNewsgroupsClustering.v2~\cite{LANG1995331}.
\end{itemize}

\section{Reasoning Tasks by Concept Weight}
\label{appendix:downstream_weight}

\begin{figure*}[h!]
  \centering
  \includegraphics[width=\linewidth]{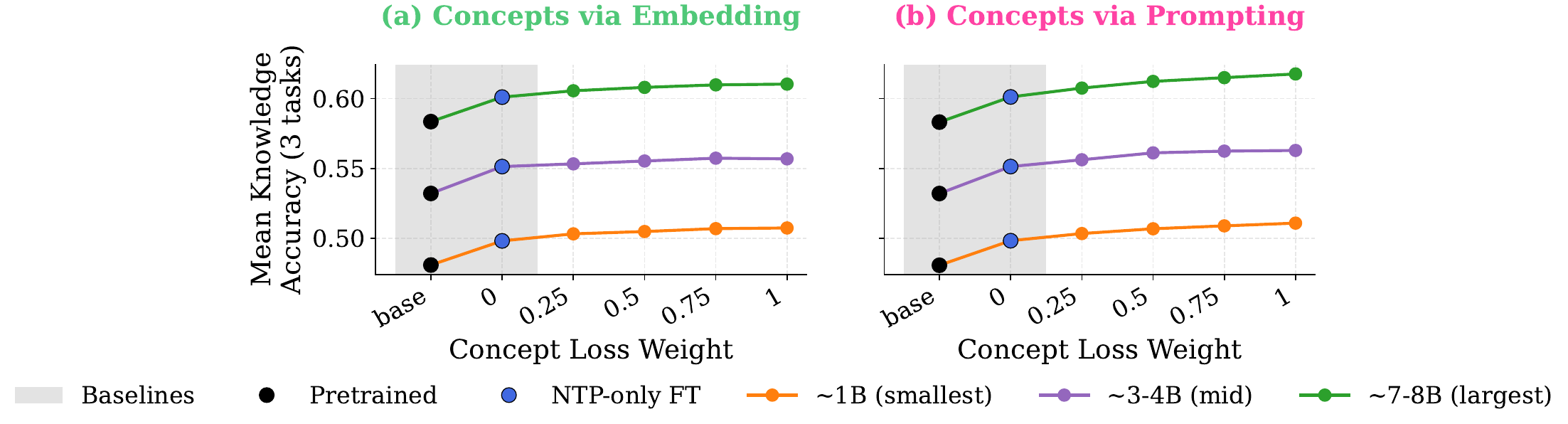}
  \caption {Knowledge reasoning accuracy increases as concept weight increases, indicating that a higher degree of concept supervision allows models to learn better factual reasoning connections. We note that the relative gains are small, but general trends hold of concept training improving performance over both the pretrained and $\lambda=0$ baselines.}
  \label{fig:reasoning_line}
\end{figure*}

We show results for the knowledge reasoning tasks separated by concept weight, aggregated over model families in Figure~\ref{fig:reasoning_line}.

\begin{figure*}[h!]
  \centering
  \includegraphics[width=\linewidth]{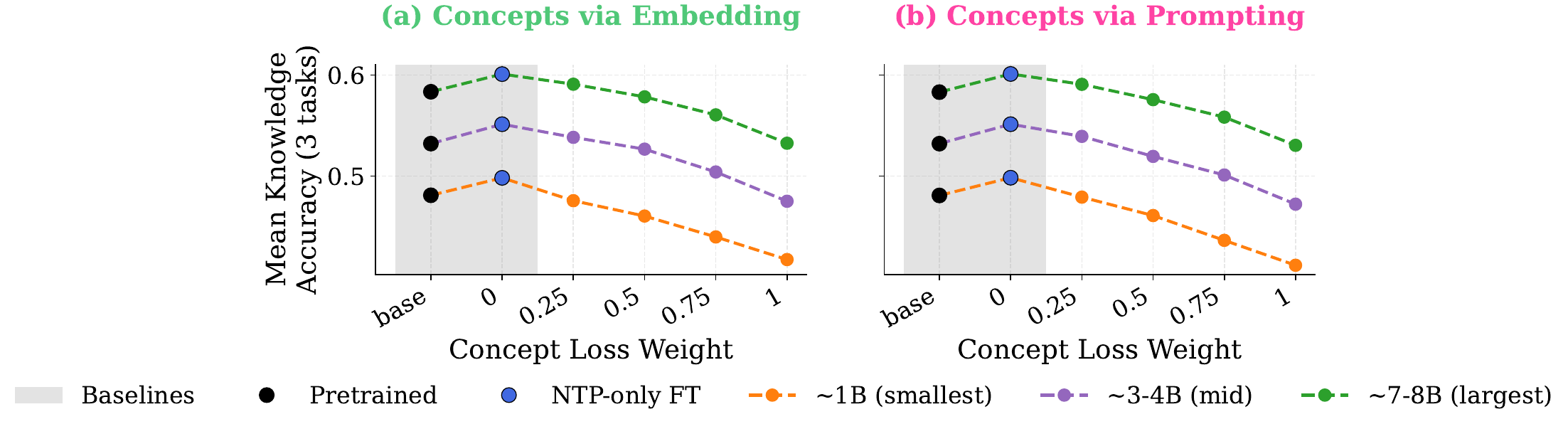}
  \caption {Knowledge reasoning accuracy decreases as concept weight increases for the models with randomized synonym sets. Without semantic groupings of concept sets, model representations become more damaged as we increase the weight on these noisy concept sets.}
  \label{fig:reasoning_line_rand}
\end{figure*}

We also show knowledge reasoning results for the randomized synonym baseline models in Figure~\ref{fig:reasoning_line_rand}.

\section{Out-of-distribution NTP Perplexity}
\label{appendix:ntp_ood}

\begin{figure*}[h!]
  \centering
  \includegraphics[width=\linewidth]{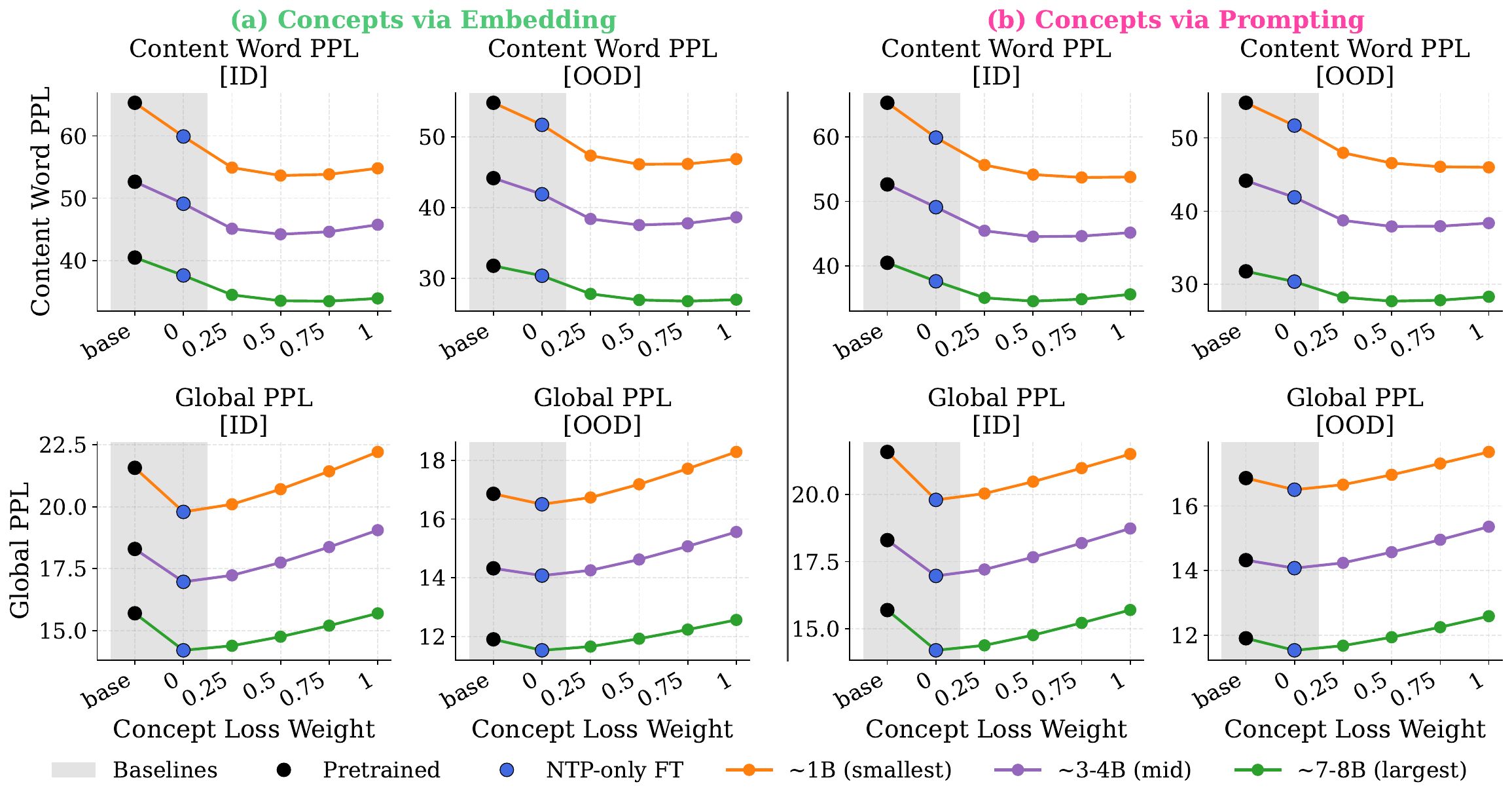}
  \caption {Concept-trained models achieve substantially lower content word perplexity than both the pretrained and NTP-only baselines both ID and OOD, while global perplexity degrades compared to the NTP-only baseline but performs either on par (ID) or slightly worse (OOD) compared to the pretrained models. Results are for models trained on C4, averaged over three training seeds and across the three model families, grouped by model size. We omit error bars because absolute perplexity values differ substantially across families due to differences in their base models.}
  \label{fig:ntp_ood}
\end{figure*}

We report both ID and OOD results together for NTP perplexity, aggregated over all model families, in Figure~\ref{fig:ntp_ood}.

\section{NTP Perplexity by Model Family}
\label{appendix:ppl_by_family}
\begin{figure*}[h!]
  \centering
  \includegraphics[width=\linewidth]{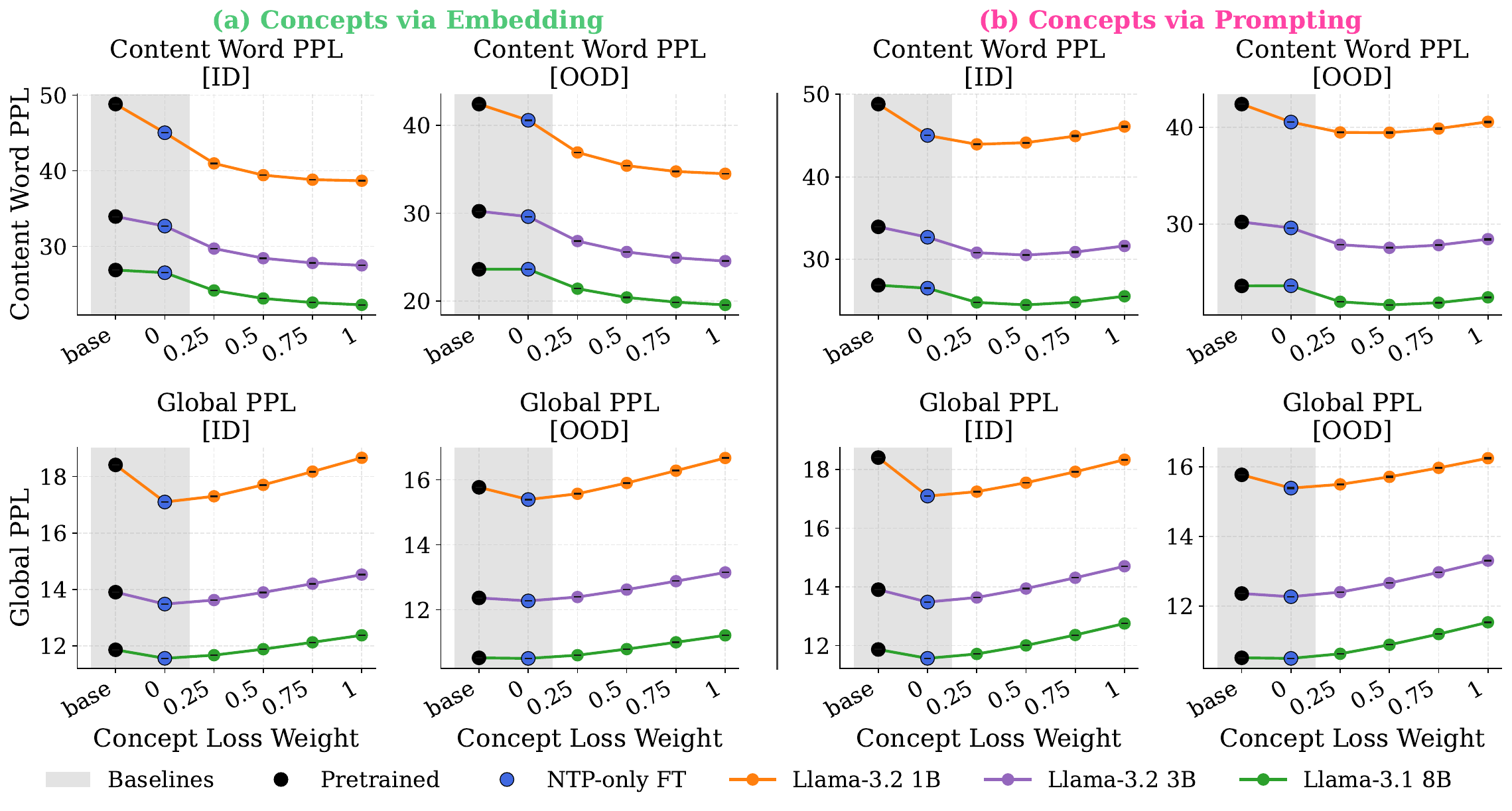}
  \caption {These plots show content word perplexity and global perplexity results for the Llama family, for models trained on C4; error bars show the standard deviation across three training seeds. Content word perplexity improves both ID and OOD compared to both baselines. Global ID perplexity at higher concept weights is comparable to the pretrained baseline but 1-2 points higher compared to the NTP-only baseline. Global OOD perplexity monotonically increases as concept weight increases, but the rise is small compared to the drop in content word perplexity (note the different y-axis scales). The prompting models appear to plateau for content word perplexity earlier than the embedding models.}
  \label{fig:llama_ppl}
\end{figure*}

\begin{figure*}[h!]
  \centering
  \includegraphics[width=\linewidth]{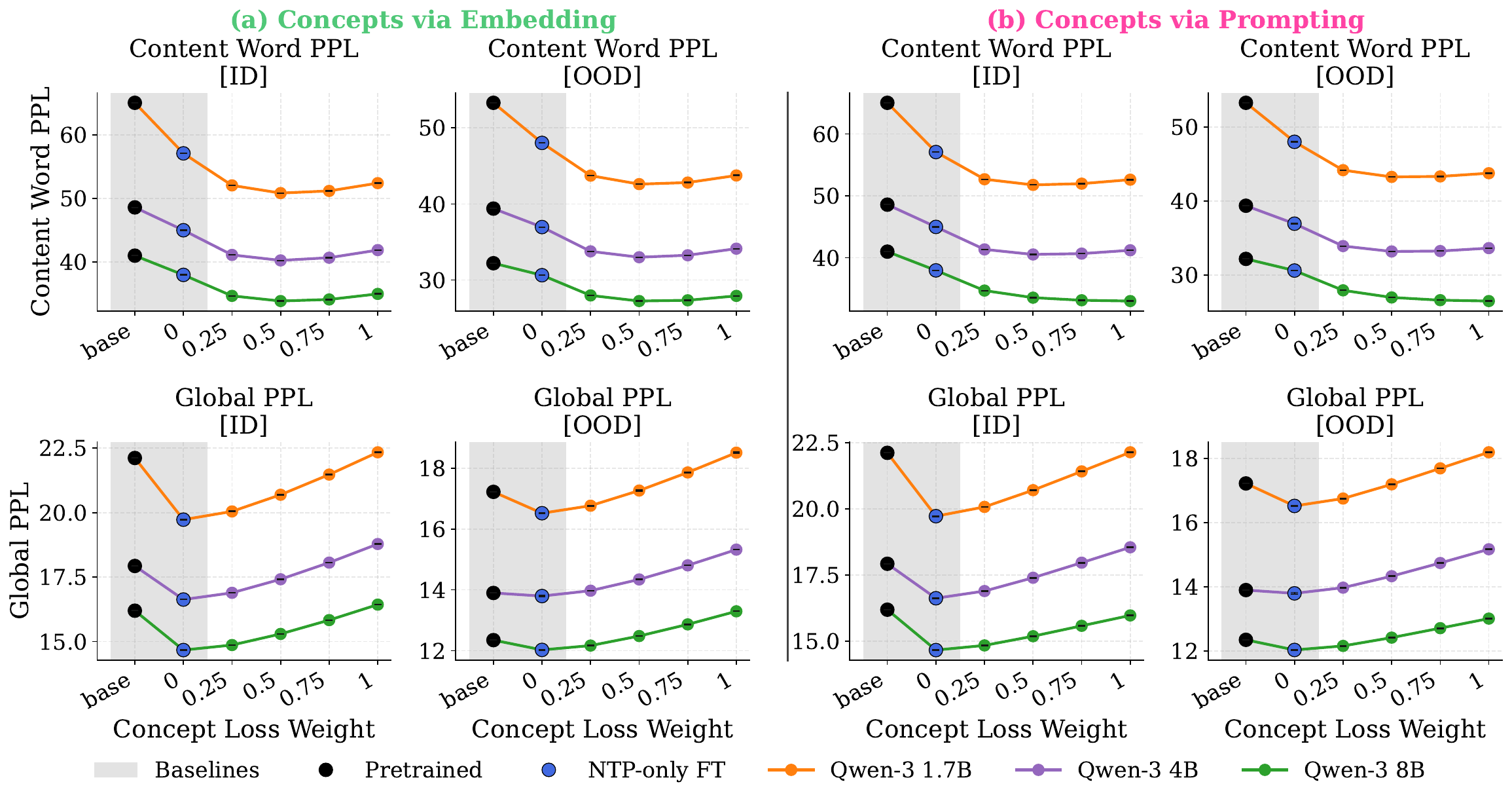}
  \caption {These plots show content word perplexity and global perplexity results for the Qwen family, for models trained on C4; error bars show the standard deviation across three training seeds. As concept weight increases, content word perplexity plateaus around $\lambda=0.5$. Similar to the Llama family, global ID perplexity at higher concept weights is comparable to the pretrained baseline but 1-2 points higher compared to the NTP-only baseline. Global OOD perplexity monotonically increases as concept weight increases, but the rise is small compared to the drop in content word perplexity (note the different y-axis scales).}
  \label{fig:qwen_ppl}
\end{figure*}

\begin{figure*}[h!]
  \centering
  \includegraphics[width=\linewidth]{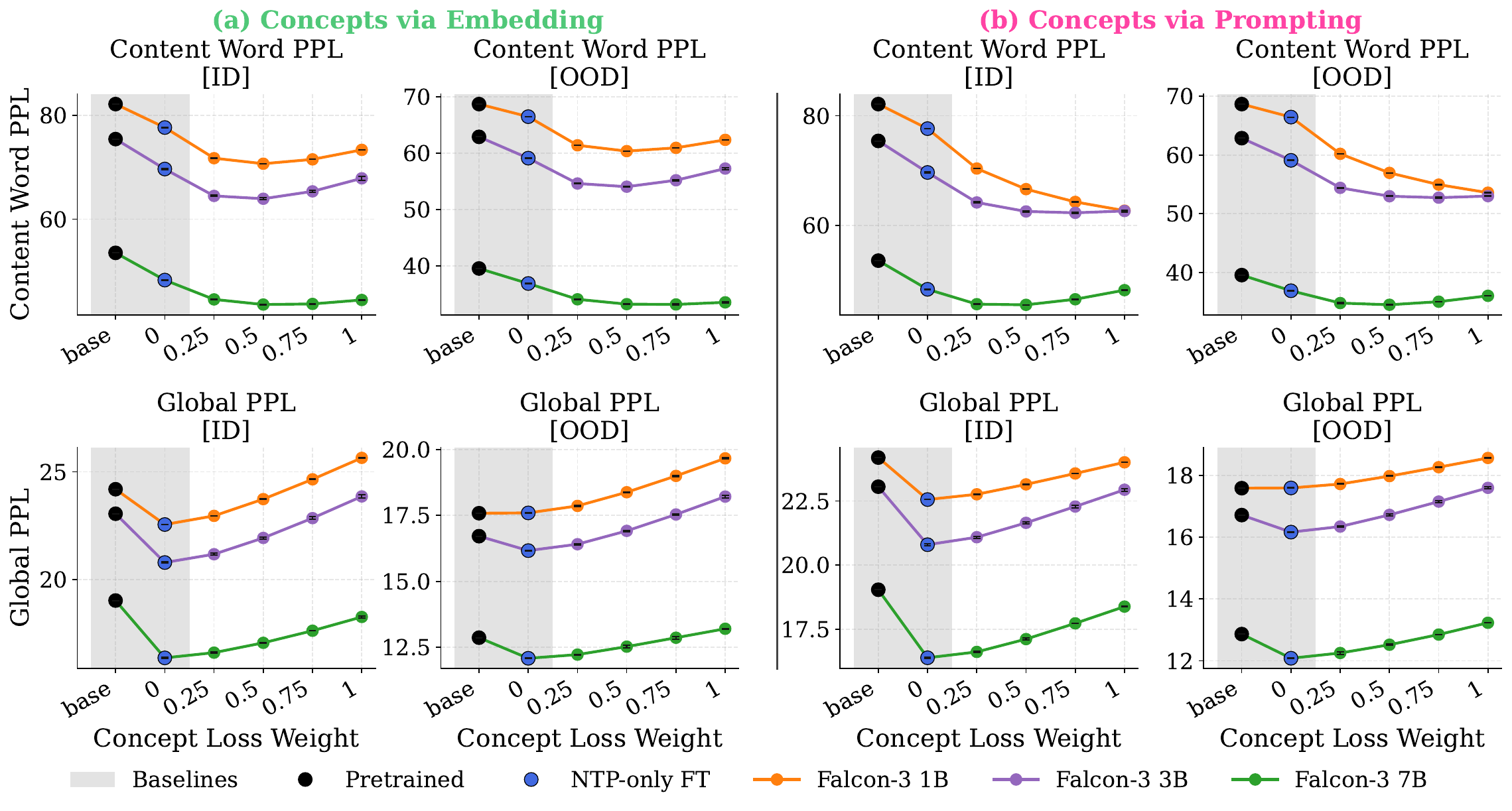}
  \caption {These plots show content word perplexity and global perplexity results for the Falcon family, for models trained on C4; error bars show the standard deviation across three training seeds. Similar to the Qwen models, the embedding models for Falcon also plateau and start degrading around concept weight $\lambda=0.5$. The prompting models continue to improve as concept supervision increases. Similar to the other model families, global ID perplexity at higher concept weights is comparable to the pretrained baseline but 1-2 points higher compared to the NTP-only baseline. Global OOD perplexity monotonically increases as concept weight increases, but the rise is small compared to the drop in content word perplexity (note the different y-axis scales).}
  \label{fig:falcon_ppl}
\end{figure*}

We show results for NTP perplexity over the Llama, Qwen, and Falcon families separately in Figure~\ref{fig:llama_ppl}, Figure~\ref{fig:qwen_ppl}, and Figure~\ref{fig:falcon_ppl} respectively.

\section{Weighted NTP Comparison by Model}
\label{appendix:wce_by_model}

\begin{figure*}[h!]
  \centering
  \includegraphics[width=\linewidth]{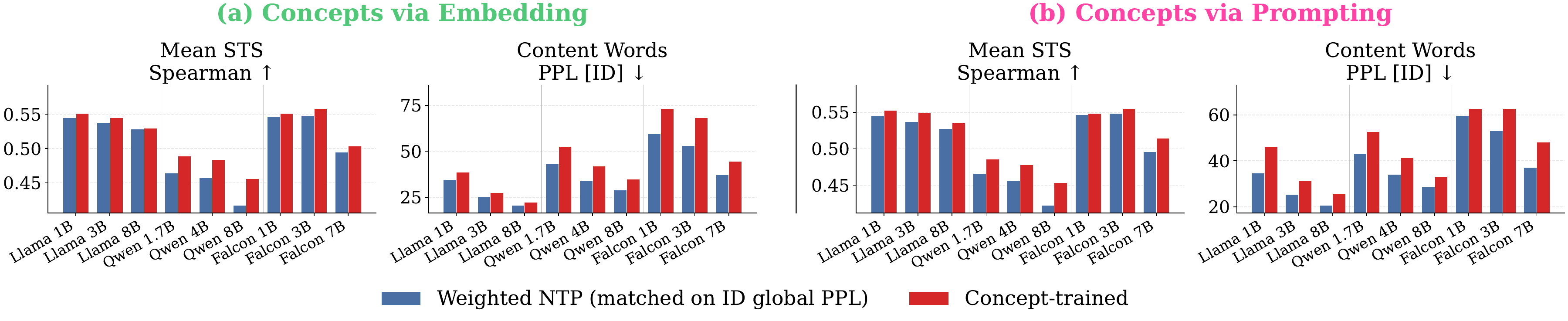}
  \caption {For every model, we see that the STS Spearman correlation for concept-trained models outperforms that of the weighted-NTP models. Conversely, the content word perplexity of the weighted-NTP models is always better than our concept-trained models, which makes sense because the weighted-NTP models specifically emphasize next token prediction at content word positions. These models are trained on C4.}
  \label{fig:wce_separate}
\end{figure*}

We show results for weighted NTP compared with concept models for each model separately in Figure~\ref{fig:wce_separate}.

\section{Saturation on Perplexity}
\label{appendix:saturation}

\begin{figure*}[h!]
  \centering
  \includegraphics[width=\linewidth]{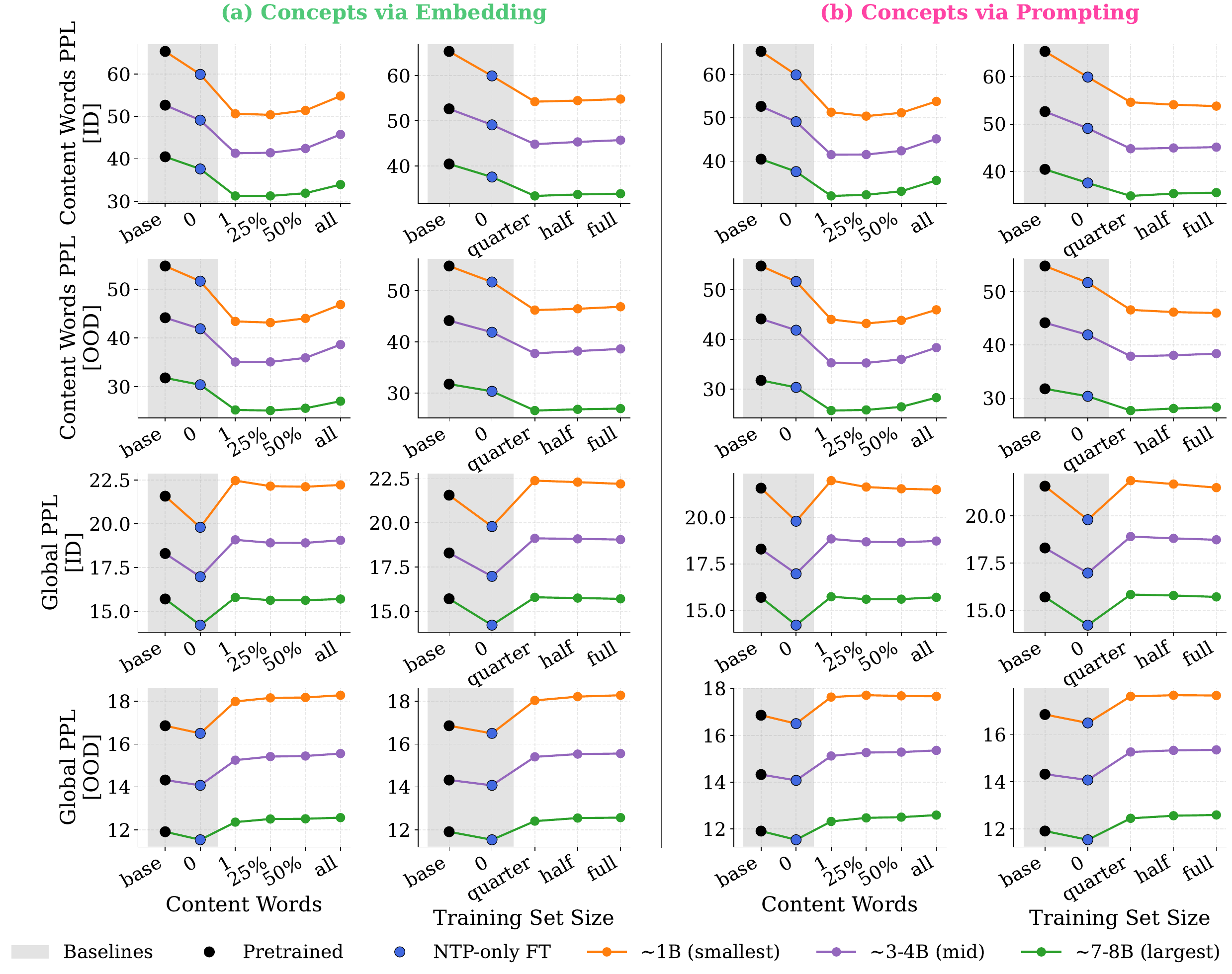}
  \caption {These plots show content word perplexity and global perplexity performance at various amounts of concept supervision. Content word ID perplexity saturates and begins overfitting when using concept supervision on only 25\% of content words, while content word OOD perplexity stays relatively constant past this point. For global perplexity, we observe that the pure NTP-trained model achieves the best performance. As the amount of concept supervision increases, the global perplexity stays relatively constant both ID and OOD for all model sizes. Results are for models trained on C4, averaged over three training seeds and aggregated over model sizes in different families.}
  \label{fig:ppl_saturation}
\end{figure*}

We observe saturation at small data amounts on both content word perplexity and global perplexity (Figure~\ref{fig:ppl_saturation}).

\section{STS Saturation by Model Family}
\label{appendix:saturation_by_family}

\begin{figure*}[h!]
  \centering
  \includegraphics[width=\linewidth]{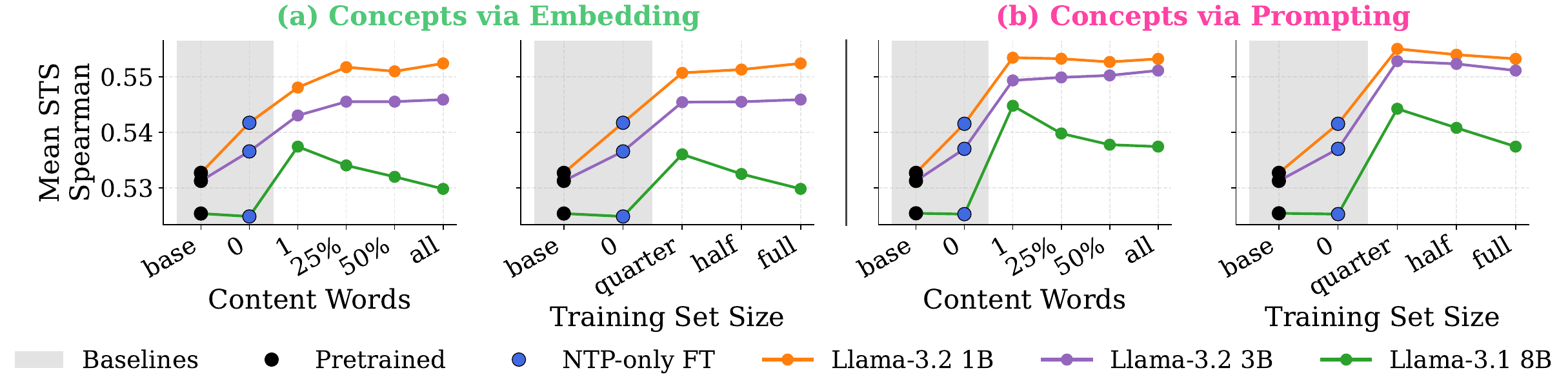}
  \caption {For Llama models, STS Spearman correlation saturates and sometimes begins overfitting after only receiving concept supervision on 1 content word per sequence, indicating that concept supervision is strong enough to affect alignment with human semantic judgments with only a sparse signal. Similarity, STS Spearman correlation also achieves its highest value with only 2,000 samples in the training set (quarter) as opposed to larger training set sizes (half, full).}
  \label{fig:llama_saturation}
\end{figure*}

\begin{figure*}[h!]
  \centering
  \includegraphics[width=\linewidth]{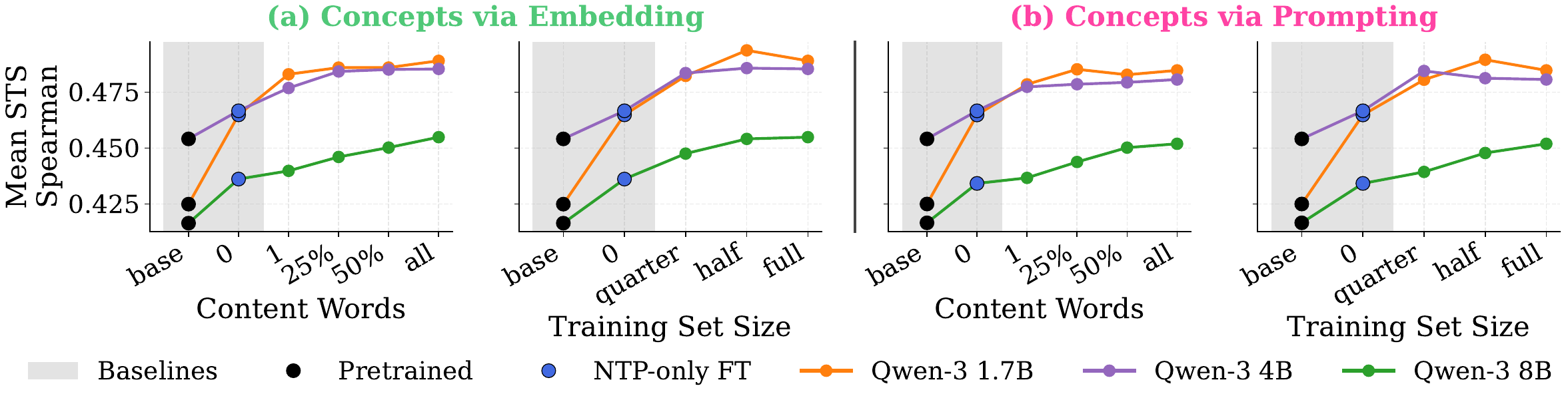}
  \caption {STS Spearman correlation for the Qwen models generally begins to plateau or improves incrementally after receiving concept supervision on 1 content word per sequence and a quarter of the entire training set size.}
  \label{fig:qwen_saturation}
\end{figure*}

\begin{figure*}[h!]
  \centering
  \includegraphics[width=\linewidth]{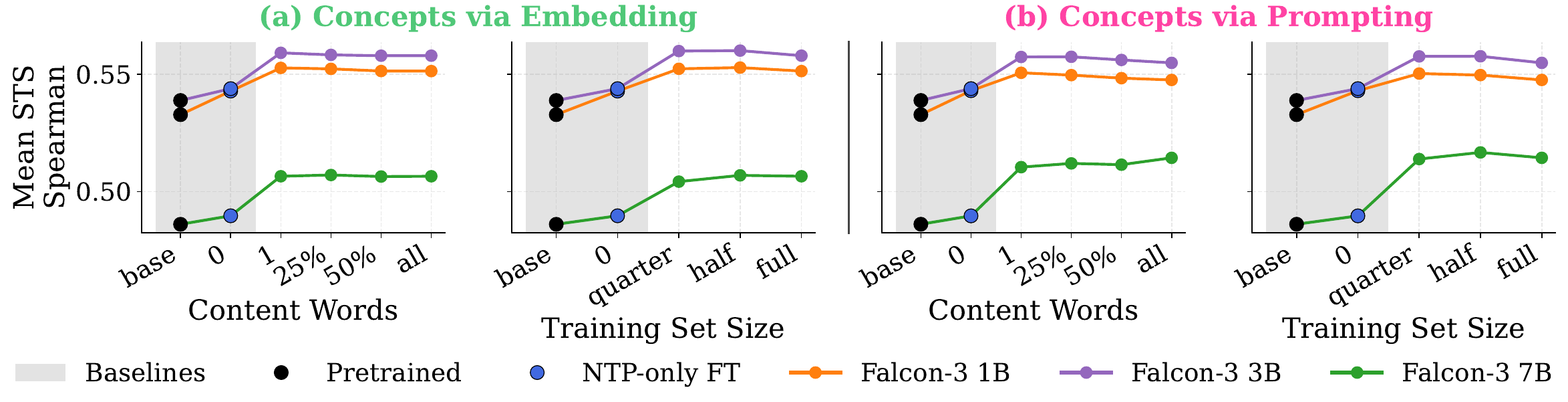}
  \caption {STS Spearman correlation for the Falcon models also achieves its highest values at 1 content word per sequence receiving concept supervision and a quarter of the training set size. At higher degrees of concept supervision, the performance slightly degrades, indicating overfitting.}
  \label{fig:falcon_saturation}
\end{figure*}

We report results per model family for STS Spearman correlation as we vary the amount of content words that receive concept supervision and the amount of training data. The plots for Llama, Qwen, and Falcon are in Figures~\ref{fig:llama_saturation},~\ref{fig:qwen_saturation}, and~\ref{fig:falcon_saturation} respectively.

\section{Contrastive Finetuning Dominates Concept Learning}
\label{appendix:contrastive}

Sections~\ref{sec:similarity} and \ref{sec:mteb} evaluate our models without any task-specific embedding finetuning, which is why our concept-trained models do not reach state-of-the-art embedding performance on STS and MTEB. In this section, we further finetune our concept-trained models using contrastive learning, which is used for many state-of-the-art embedding models~\cite{lee2025geminiembeddinggeneralizableembeddings, qwen3embedding, lee2025nvembed}, to investigate whether our models can achieve better performance using this method.

\paragraph{Experimental Setup.} We freeze each post-trained model and train an identically initialized LoRA on top of it with a SimCSE-style supervised contrastive objective~\cite{gao-etal-2021-simcse}. The contrastive training data is a 50{,}000-triplet subsample of the SNLI and MNLI (anchor, entailment, contradiction) triplets~\cite{bowman-etal-2015-large, williams-etal-2018-broad}. As throughout the paper, $\lambda$ denotes the concept weight used to train the initial adapter.

We run our experiment on all 9 models as reported in the paper (3 sizes for each model family). We compare five types within each model: pretrained, NTP-only ($\lambda=0$), concept ($\lambda=1$), randomized synonym with $\lambda=0.25$, and randomized synonym with $\lambda=1$. We chose the first two as baselines, the third as our presented model, and the last two as intentionally degraded baselines for comparison.


\paragraph{Results.} Tables~\ref{tab:contrastive_sts_c4} and \ref{tab:contrastive_short_c4} report every C4-trained model's score on STS and MTEB short tasks before and after contrastive finetuning; Tables~\ref{tab:contrastive_sts_owt} and \ref{tab:contrastive_short_owt} do the same for the OpenWebText-trained models. Before contrastive finetuning, the five types within a single row differ by as much as 30.0 points on C4 and 20.4 points on OpenWebText. Afterwards, they agree to within 1.16 points on C4 and 1.73 points on OpenWebText. These results suggest that contrastive finetuning dominates over our post-training scheme in general, which aligns with our intuition because contrastive learning directly trains models' embedding behavior. In contrast, our concept learning technique is a variant on vanilla NTP that improves embedding performance as a byproduct. The signal from concept training is therefore unsurprisingly dominated here: contrastive finetuning has so large an effect that this experiment cannot distinguish between our baselines, degraded models, and concept-trained models.


\clearpage


\begin{table*}[h]
  \centering
  \small
  \begin{tabular}{lcc}
  \toprule
  \textbf{Model family} & \textbf{C4} & \textbf{OpenWebText} \\
  \midrule
  Llama 3 & $46.7 \pm 22.0$ & $58.6 \pm 12.1$ \\
  Qwen 3 & $46.1 \pm 21.9$ & $57.8 \pm 12.2$ \\
  Falcon 3 & $47.8 \pm 22.4$ & $58.2 \pm 12.3$ \\
  \bottomrule
  \end{tabular}
  \caption{Mean number of content words (complete-word single-token nouns, verbs, and adjectives) per sequence, $\pm$ one standard deviation. This stage depends only on the tokenizer, so it is identical across the three sizes within each model family.}
  \label{tab:content_words}
  \end{table*}
  
  \begin{table*}[h!]
  \centering
  \small
  \begin{tabular}{lccccc}
  \toprule
  & & \multicolumn{2}{c}{\textcolor{mygreen}{\textbf{Embedding}}} & \multicolumn{2}{c}{\textcolor{WildStrawberry}{\textbf{Prompting}}} \\
  \cmidrule(lr){3-4} \cmidrule(lr){5-6}
  \textbf{Model} & \textbf{Candidates} & \textbf{Synonyms} & \textbf{Coverage} & \textbf{Synonyms} & \textbf{Coverage} \\
  \midrule
  \multicolumn{6}{l}{\textit{C4}} \\
  \quad Llama 3.2 1B & $92.5 \pm 10.9$ & $7.06 \pm 3.74$ & 88.3\% & $6.69 \pm 3.59$ & 87.2\% \\
  \quad Llama 3.2 3B & $91.5 \pm 11.4$ & $6.47 \pm 3.94$ & 85.7\% & $5.52 \pm 3.36$ & 92.4\% \\
  \quad Llama 3.1 8B & $91.4 \pm 11.8$ & $6.01 \pm 4.01$ & 83.7\% & $6.41 \pm 3.16$ & 95.3\% \\
  \quad Qwen 3 1.7B & $92.7 \pm 10.8$ & $8.11 \pm 2.91$ & 96.4\% & $2.99 \pm 2.13$ & 90.5\% \\
  \quad Qwen 3 4B & $91.9 \pm 10.4$ & $8.13 \pm 2.87$ & 96.7\% & $5.39 \pm 3.66$ & 83.5\% \\
  \quad Qwen 3 8B & $92.6 \pm 10.1$ & $8.16 \pm 2.81$ & 97.1\% & $5.22 \pm 3.78$ & 76.2\% \\
  \quad Falcon 3 1B & $94.6 \pm 7.3$ & $8.04 \pm 2.94$ & 96.3\% & $2.04 \pm 3.30$ & 34.7\% \\
  \quad Falcon 3 3B & $94.1 \pm 9.0$ & $8.10 \pm 2.88$ & 96.7\% & $4.88 \pm 3.39$ & 83.6\% \\
  \quad Falcon 3 7B & $93.4 \pm 8.4$ & $8.00 \pm 2.94$ & 96.3\% & $6.49 \pm 2.98$ & 94.8\% \\
  \midrule
  \multicolumn{6}{l}{\textit{OpenWebText}} \\
  \quad Llama 3.2 1B & $92.1 \pm 10.6$ & $7.13 \pm 3.70$ & 89.8\% & $5.48 \pm 3.95$ & 76.4\% \\
  \quad Llama 3.2 3B & $90.8 \pm 11.4$ & $6.44 \pm 3.95$ & 86.2\% & $5.34 \pm 3.48$ & 87.9\% \\
  \quad Llama 3.1 8B & $91.1 \pm 11.5$ & $6.02 \pm 4.00$ & 84.5\% & $6.02 \pm 3.40$ & 90.2\% \\
  \quad Qwen 3 1.7B & $92.3 \pm 10.4$ & $8.21 \pm 2.86$ & 96.8\% & $2.61 \pm 2.03$ & 83.3\% \\
  \quad Qwen 3 4B & $91.4 \pm 10.2$ & $8.20 \pm 2.84$ & 97.0\% & $5.13 \pm 3.70$ & 80.7\% \\
  \quad Qwen 3 8B & $92.3 \pm 9.9$ & $8.25 \pm 2.77$ & 97.4\% & $5.09 \pm 3.84$ & 73.5\% \\
  \quad Falcon 3 1B & $94.8 \pm 6.8$ & $8.12 \pm 2.93$ & 96.3\% & $1.72 \pm 3.06$ & 31.0\% \\
  \quad Falcon 3 3B & $94.4 \pm 8.3$ & $8.20 \pm 2.85$ & 96.8\% & $4.42 \pm 3.35$ & 79.9\% \\
  \quad Falcon 3 7B & $93.7 \pm 8.0$ & $8.09 \pm 2.92$ & 96.5\% & $6.04 \pm 3.27$ & 89.9\% \\
  \bottomrule
  \end{tabular}
  \caption{Concept set sizes per content word, $\pm$ one standard deviation. \textbf{Candidates} is the size of the shared candidate pool (top 100 next-token predictions after filtering) before applying either the embedding or prompting methods. \textbf{Synonyms} is the mean size of the final concept set under each method, averaged over all content words including those that receive no synonyms. \textbf{Coverage} is the fraction of content words whose concept set is non-empty.}
  \label{tab:concept_sizes}
  \end{table*}

\begin{table*}[h!]
\centering
\footnotesize
\begin{tabular}{llccccc}
\toprule
\cmod{\textbf{Model}} & \cstg{} & \cnum{\textbf{Pretrained}} & \cnum{\textbf{NTP-only}} & \cnum{\textbf{Concept}} & \cnum{\textbf{\makecell{Rand.\\$\lambda{=}0.25$}}} & \cnum{\textbf{\makecell{Rand.\\$\lambda{=}1$}}} \\
\midrule
\multicolumn{7}{l}{\textit{Concepts via Embedding}} \\
\midrule
\multirow{2}{*}{\cmod{Llama 1B}} & \cstg{Before} & \cnum{53.27} & \cnum{54.16} & \cnum{55.13} & \cnum{52.01} & \cnum{46.77} \\
\cmod{} & \cstg{After} & \cnum{80.49} & \cnum{80.48} & \cnum{80.52} & \cnum{80.49} & \cnum{80.39} \\
\dashsep{7}
\multirow{2}{*}{\cmod{Llama 3B}} & \cstg{Before} & \cnum{53.13} & \cnum{53.54} & \cnum{54.47} & \cnum{50.50} & \cnum{45.23} \\
\cmod{} & \cstg{After} & \cnum{81.49} & \cnum{81.35} & \cnum{81.37} & \cnum{81.31} & \cnum{81.35} \\
\dashsep{7}
\multirow{2}{*}{\cmod{Llama 8B}} & \cstg{Before} & \cnum{52.54} & \cnum{52.49} & \cnum{52.94} & \cnum{48.61} & \cnum{44.19} \\
\cmod{} & \cstg{After} & \cnum{82.41} & \cnum{82.23} & \cnum{82.44} & \cnum{82.25} & \cnum{82.01} \\
\dashsep{7}
\multirow{2}{*}{\cmod{Qwen 1.7B}} & \cstg{Before} & \cnum{42.50} & \cnum{44.19} & \cnum{48.49} & \cnum{42.45} & \cnum{18.94} \\
\cmod{} & \cstg{After} & \cnum{79.64} & \cnum{80.05} & \cnum{79.77} & \cnum{79.23} & \cnum{79.90} \\
\dashsep{7}
\multirow{2}{*}{\cmod{Qwen 4B}} & \cstg{Before} & \cnum{45.41} & \cnum{46.42} & \cnum{48.32} & \cnum{41.27} & \cnum{35.61} \\
\cmod{} & \cstg{After} & \cnum{81.12} & \cnum{81.24} & \cnum{81.00} & \cnum{81.24} & \cnum{80.95} \\
\dashsep{7}
\multirow{2}{*}{\cmod{Qwen 8B}} & \cstg{Before} & \cnum{41.64} & \cnum{43.71} & \cnum{45.56} & \cnum{37.54} & \cnum{33.38} \\
\cmod{} & \cstg{After} & \cnum{82.15} & \cnum{81.30} & \cnum{81.45} & \cnum{81.49} & \cnum{81.63} \\
\dashsep{7}
\multirow{2}{*}{\cmod{Falcon 1B}} & \cstg{Before} & \cnum{53.28} & \cnum{54.35} & \cnum{55.11} & \cnum{54.04} & \cnum{50.61} \\
\cmod{} & \cstg{After} & \cnum{75.85} & \cnum{75.09} & \cnum{75.31} & \cnum{75.09} & \cnum{75.44} \\
\dashsep{7}
\multirow{2}{*}{\cmod{Falcon 3B}} & \cstg{Before} & \cnum{53.89} & \cnum{54.38} & \cnum{55.86} & \cnum{52.84} & \cnum{46.66} \\
\cmod{} & \cstg{After} & \cnum{77.59} & \cnum{76.73} & \cnum{77.16} & \cnum{77.07} & \cnum{76.87} \\
\dashsep{7}
\multirow{2}{*}{\cmod{Falcon 7B}} & \cstg{Before} & \cnum{48.61} & \cnum{49.01} & \cnum{50.37} & \cnum{45.45} & \cnum{41.69} \\
\cmod{} & \cstg{After} & \cnum{80.50} & \cnum{79.80} & \cnum{80.23} & \cnum{80.34} & \cnum{80.25} \\
\midrule
\multicolumn{7}{l}{\textit{Concepts via Prompting}} \\
\midrule
\multirow{2}{*}{\cmod{Llama 1B}} & \cstg{Before} & \cnum{53.27} & \cnum{54.11} & \cnum{55.25} & \cnum{52.80} & \cnum{47.67} \\
\cmod{} & \cstg{After} & \cnum{80.49} & \cnum{80.50} & \cnum{80.52} & \cnum{80.39} & \cnum{80.68} \\
\dashsep{7}
\multirow{2}{*}{\cmod{Llama 3B}} & \cstg{Before} & \cnum{53.13} & \cnum{53.54} & \cnum{54.92} & \cnum{50.81} & \cnum{45.81} \\
\cmod{} & \cstg{After} & \cnum{81.49} & \cnum{81.36} & \cnum{81.45} & \cnum{81.11} & \cnum{81.19} \\
\dashsep{7}
\multirow{2}{*}{\cmod{Llama 8B}} & \cstg{Before} & \cnum{52.54} & \cnum{52.48} & \cnum{53.51} & \cnum{49.23} & \cnum{45.10} \\
\cmod{} & \cstg{After} & \cnum{82.41} & \cnum{82.37} & \cnum{82.24} & \cnum{82.01} & \cnum{81.94} \\
\dashsep{7}
\multirow{2}{*}{\cmod{Qwen 1.7B}} & \cstg{Before} & \cnum{42.50} & \cnum{44.19} & \cnum{48.18} & \cnum{42.39} & \cnum{18.19} \\
\cmod{} & \cstg{After} & \cnum{79.64} & \cnum{80.02} & \cnum{79.93} & \cnum{80.04} & \cnum{80.07} \\
\dashsep{7}
\multirow{2}{*}{\cmod{Qwen 4B}} & \cstg{Before} & \cnum{45.41} & \cnum{46.42} & \cnum{47.80} & \cnum{40.10} & \cnum{33.74} \\
\cmod{} & \cstg{After} & \cnum{81.12} & \cnum{81.24} & \cnum{81.00} & \cnum{81.51} & \cnum{80.85} \\
\dashsep{7}
\multirow{2}{*}{\cmod{Qwen 8B}} & \cstg{Before} & \cnum{41.65} & \cnum{43.13} & \cnum{45.39} & \cnum{35.64} & \cnum{30.18} \\
\cmod{} & \cstg{After} & \cnum{81.72} & \cnum{80.94} & \cnum{81.46} & \cnum{81.06} & \cnum{81.34} \\
\dashsep{7}
\multirow{2}{*}{\cmod{Falcon 1B}} & \cstg{Before} & \cnum{53.28} & \cnum{54.34} & \cnum{54.81} & \cnum{53.83} & \cnum{47.83} \\
\cmod{} & \cstg{After} & \cnum{75.79} & \cnum{75.09} & \cnum{75.28} & \cnum{75.36} & \cnum{75.43} \\
\dashsep{7}
\multirow{2}{*}{\cmod{Falcon 3B}} & \cstg{Before} & \cnum{53.89} & \cnum{54.39} & \cnum{55.52} & \cnum{52.68} & \cnum{47.74} \\
\cmod{} & \cstg{After} & \cnum{77.59} & \cnum{76.92} & \cnum{77.33} & \cnum{76.94} & \cnum{76.79} \\
\dashsep{7}
\multirow{2}{*}{\cmod{Falcon 7B}} & \cstg{Before} & \cnum{48.61} & \cnum{49.02} & \cnum{51.42} & \cnum{45.21} & \cnum{41.11} \\
\cmod{} & \cstg{After} & \cnum{80.50} & \cnum{79.72} & \cnum{79.97} & \cnum{80.32} & \cnum{80.25} \\
\bottomrule
\end{tabular}
\caption{STS (9 tasks) scores ($\times 100$) for every C4-trained model, before and after identical SimCSE-style contrastive finetuning. \textbf{Before} rows are the post-trained models the rest of the paper evaluates; \textbf{After} rows are those same models after contrastive finetuning. Before finetuning the five types within each model differ by as much as 30.0 points within a single row; afterwards they agree to within 0.86 points everywhere, including the randomized-synonym variants trained on deliberately corrupted supervision.}
\label{tab:contrastive_sts_c4}
\end{table*}

\begin{table*}[h!]
\centering
\footnotesize
\begin{tabular}{llccccc}
\toprule
\cmod{\textbf{Model}} & \cstg{} & \cnum{\textbf{Pretrained}} & \cnum{\textbf{NTP-only}} & \cnum{\textbf{Concept}} & \cnum{\textbf{\makecell{Rand.\\$\lambda{=}0.25$}}} & \cnum{\textbf{\makecell{Rand.\\$\lambda{=}1$}}} \\
\midrule
\multicolumn{7}{l}{\textit{Concepts via Embedding}} \\
\midrule
\multirow{2}{*}{\cmod{Llama 1B}} & \cstg{Before} & \cnum{47.08} & \cnum{47.56} & \cnum{49.11} & \cnum{45.64} & \cnum{44.18} \\
\cmod{} & \cstg{After} & \cnum{66.05} & \cnum{65.70} & \cnum{65.93} & \cnum{66.23} & \cnum{66.12} \\
\dashsep{7}
\multirow{2}{*}{\cmod{Llama 3B}} & \cstg{Before} & \cnum{48.38} & \cnum{48.55} & \cnum{49.49} & \cnum{45.50} & \cnum{43.95} \\
\cmod{} & \cstg{After} & \cnum{66.30} & \cnum{66.66} & \cnum{66.40} & \cnum{66.79} & \cnum{67.45} \\
\dashsep{7}
\multirow{2}{*}{\cmod{Llama 8B}} & \cstg{Before} & \cnum{49.12} & \cnum{49.36} & \cnum{49.82} & \cnum{44.98} & \cnum{40.32} \\
\cmod{} & \cstg{After} & \cnum{65.33} & \cnum{65.56} & \cnum{65.76} & \cnum{65.92} & \cnum{65.86} \\
\dashsep{7}
\multirow{2}{*}{\cmod{Qwen 1.7B}} & \cstg{Before} & \cnum{37.28} & \cnum{37.88} & \cnum{41.92} & \cnum{37.24} & \cnum{25.53} \\
\cmod{} & \cstg{After} & \cnum{64.38} & \cnum{64.37} & \cnum{64.28} & \cnum{63.93} & \cnum{64.65} \\
\dashsep{7}
\multirow{2}{*}{\cmod{Qwen 4B}} & \cstg{Before} & \cnum{39.30} & \cnum{40.14} & \cnum{42.10} & \cnum{36.66} & \cnum{33.70} \\
\cmod{} & \cstg{After} & \cnum{65.10} & \cnum{65.08} & \cnum{65.36} & \cnum{64.91} & \cnum{64.81} \\
\dashsep{7}
\multirow{2}{*}{\cmod{Qwen 8B}} & \cstg{Before} & \cnum{35.04} & \cnum{36.78} & \cnum{38.01} & \cnum{33.49} & \cnum{31.41} \\
\cmod{} & \cstg{After} & \cnum{65.39} & \cnum{65.67} & \cnum{65.74} & \cnum{65.25} & \cnum{65.05} \\
\dashsep{7}
\multirow{2}{*}{\cmod{Falcon 1B}} & \cstg{Before} & \cnum{47.40} & \cnum{49.14} & \cnum{49.95} & \cnum{48.38} & \cnum{46.19} \\
\cmod{} & \cstg{After} & \cnum{63.35} & \cnum{63.02} & \cnum{62.90} & \cnum{62.82} & \cnum{63.21} \\
\dashsep{7}
\multirow{2}{*}{\cmod{Falcon 3B}} & \cstg{Before} & \cnum{45.88} & \cnum{48.07} & \cnum{48.61} & \cnum{46.43} & \cnum{42.62} \\
\cmod{} & \cstg{After} & \cnum{64.13} & \cnum{63.69} & \cnum{64.05} & \cnum{64.02} & \cnum{64.32} \\
\dashsep{7}
\multirow{2}{*}{\cmod{Falcon 7B}} & \cstg{Before} & \cnum{43.39} & \cnum{44.02} & \cnum{44.96} & \cnum{41.33} & \cnum{38.95} \\
\cmod{} & \cstg{After} & \cnum{64.10} & \cnum{64.22} & \cnum{64.37} & \cnum{64.02} & \cnum{64.30} \\
\midrule
\multicolumn{7}{l}{\textit{Concepts via Prompting}} \\
\midrule
\multirow{2}{*}{\cmod{Llama 1B}} & \cstg{Before} & \cnum{47.08} & \cnum{47.50} & \cnum{49.09} & \cnum{46.71} & \cnum{45.60} \\
\cmod{} & \cstg{After} & \cnum{66.04} & \cnum{65.83} & \cnum{65.92} & \cnum{65.85} & \cnum{66.02} \\
\dashsep{7}
\multirow{2}{*}{\cmod{Llama 3B}} & \cstg{Before} & \cnum{48.38} & \cnum{48.50} & \cnum{49.58} & \cnum{45.80} & \cnum{44.31} \\
\cmod{} & \cstg{After} & \cnum{66.24} & \cnum{66.71} & \cnum{66.44} & \cnum{66.69} & \cnum{66.60} \\
\dashsep{7}
\multirow{2}{*}{\cmod{Llama 8B}} & \cstg{Before} & \cnum{49.11} & \cnum{49.15} & \cnum{50.16} & \cnum{45.92} & \cnum{41.68} \\
\cmod{} & \cstg{After} & \cnum{65.33} & \cnum{65.74} & \cnum{65.85} & \cnum{65.70} & \cnum{65.56} \\
\dashsep{7}
\multirow{2}{*}{\cmod{Qwen 1.7B}} & \cstg{Before} & \cnum{37.29} & \cnum{37.88} & \cnum{41.61} & \cnum{37.37} & \cnum{23.65} \\
\cmod{} & \cstg{After} & \cnum{64.38} & \cnum{64.36} & \cnum{64.31} & \cnum{64.76} & \cnum{64.87} \\
\dashsep{7}
\multirow{2}{*}{\cmod{Qwen 4B}} & \cstg{Before} & \cnum{39.29} & \cnum{40.18} & \cnum{40.95} & \cnum{36.00} & \cnum{32.57} \\
\cmod{} & \cstg{After} & \cnum{65.10} & \cnum{65.08} & \cnum{65.05} & \cnum{64.85} & \cnum{64.93} \\
\dashsep{7}
\multirow{2}{*}{\cmod{Qwen 8B}} & \cstg{Before} & \cnum{35.06} & \cnum{36.48} & \cnum{37.36} & \cnum{32.51} & \cnum{29.85} \\
\cmod{} & \cstg{After} & \cnum{65.69} & \cnum{65.55} & \cnum{65.30} & \cnum{65.77} & \cnum{65.76} \\
\dashsep{7}
\multirow{2}{*}{\cmod{Falcon 1B}} & \cstg{Before} & \cnum{47.40} & \cnum{49.13} & \cnum{49.36} & \cnum{48.12} & \cnum{44.10} \\
\cmod{} & \cstg{After} & \cnum{62.72} & \cnum{62.52} & \cnum{62.47} & \cnum{62.49} & \cnum{62.72} \\
\dashsep{7}
\multirow{2}{*}{\cmod{Falcon 3B}} & \cstg{Before} & \cnum{45.88} & \cnum{48.00} & \cnum{48.18} & \cnum{46.27} & \cnum{44.63} \\
\cmod{} & \cstg{After} & \cnum{64.13} & \cnum{63.96} & \cnum{63.97} & \cnum{64.06} & \cnum{64.17} \\
\dashsep{7}
\multirow{2}{*}{\cmod{Falcon 7B}} & \cstg{Before} & \cnum{43.39} & \cnum{44.08} & \cnum{45.86} & \cnum{41.42} & \cnum{38.63} \\
\cmod{} & \cstg{After} & \cnum{64.10} & \cnum{64.05} & \cnum{64.24} & \cnum{63.99} & \cnum{64.09} \\
\bottomrule
\end{tabular}
\caption{Short-context MTEB (15 tasks) scores ($\times 100$) for every C4-trained model, before and after identical SimCSE-style contrastive finetuning (C4). \textbf{Before} rows are the post-trained models the rest of the paper evaluates; \textbf{After} rows are those same models after contrastive finetuning. Before finetuning the five types within each model differ by as much as 18.0 points within a single row; afterwards they agree to within 1.16 points everywhere, including the randomized-synonym variants trained on deliberately corrupted supervision.}

\label{tab:contrastive_short_c4}
\end{table*}

\begin{table*}[h!]
\centering
\footnotesize
\begin{tabular}{llccccc}
\toprule
\cmod{\textbf{Model}} & \cstg{} & \cnum{\textbf{Pretrained}} & \cnum{\textbf{NTP-only}} & \cnum{\textbf{Concept}} & \cnum{\textbf{\makecell{Rand.\\$\lambda{=}0.25$}}} & \cnum{\textbf{\makecell{Rand.\\$\lambda{=}1$}}} \\
\midrule
\multicolumn{7}{l}{\textit{Concepts via Embedding}} \\
\midrule
\multirow{2}{*}{\cmod{Llama 1B}} & \cstg{Before} & \cnum{53.27} & \cnum{53.20} & \cnum{54.15} & \cnum{51.21} & \cnum{45.83} \\
\cmod{} & \cstg{After} & \cnum{80.44} & \cnum{80.45} & \cnum{80.36} & \cnum{80.43} & \cnum{80.48} \\
\dashsep{7}
\multirow{2}{*}{\cmod{Llama 3B}} & \cstg{Before} & \cnum{53.13} & \cnum{53.24} & \cnum{53.70} & \cnum{50.16} & \cnum{45.94} \\
\cmod{} & \cstg{After} & \cnum{81.49} & \cnum{81.43} & \cnum{81.47} & \cnum{81.18} & \cnum{81.32} \\
\dashsep{7}
\multirow{2}{*}{\cmod{Llama 8B}} & \cstg{Before} & \cnum{52.54} & \cnum{52.44} & \cnum{52.99} & \cnum{49.43} & \cnum{45.26} \\
\cmod{} & \cstg{After} & \cnum{82.41} & \cnum{82.22} & \cnum{82.46} & \cnum{82.25} & \cnum{82.31} \\
\dashsep{7}
\multirow{2}{*}{\cmod{Qwen 1.7B}} & \cstg{Before} & \cnum{42.50} & \cnum{44.89} & \cnum{46.30} & \cnum{41.08} & \cnum{35.05} \\
\cmod{} & \cstg{After} & \cnum{79.64} & \cnum{79.59} & \cnum{79.74} & \cnum{80.16} & \cnum{79.76} \\
\dashsep{7}
\multirow{2}{*}{\cmod{Qwen 4B}} & \cstg{Before} & \cnum{45.42} & \cnum{45.10} & \cnum{47.16} & \cnum{40.92} & \cnum{35.94} \\
\cmod{} & \cstg{After} & \cnum{81.12} & \cnum{80.93} & \cnum{80.84} & \cnum{80.71} & \cnum{81.52} \\
\dashsep{7}
\multirow{2}{*}{\cmod{Qwen 8B}} & \cstg{Before} & \cnum{41.65} & \cnum{30.27} & \cnum{43.42} & \cnum{29.33} & \cnum{23.04} \\
\cmod{} & \cstg{After} & \cnum{81.72} & \cnum{81.21} & \cnum{81.18} & \cnum{80.97} & \cnum{81.07} \\
\dashsep{7}
\multirow{2}{*}{\cmod{Falcon 1B}} & \cstg{Before} & \cnum{53.28} & \cnum{53.10} & \cnum{54.52} & \cnum{53.16} & \cnum{48.59} \\
\cmod{} & \cstg{After} & \cnum{75.83} & \cnum{75.12} & \cnum{75.37} & \cnum{75.25} & \cnum{75.05} \\
\dashsep{7}
\multirow{2}{*}{\cmod{Falcon 3B}} & \cstg{Before} & \cnum{53.89} & \cnum{53.11} & \cnum{54.74} & \cnum{52.25} & \cnum{44.45} \\
\cmod{} & \cstg{After} & \cnum{77.40} & \cnum{76.52} & \cnum{76.91} & \cnum{76.85} & \cnum{76.82} \\
\dashsep{7}
\multirow{2}{*}{\cmod{Falcon 7B}} & \cstg{Before} & \cnum{48.60} & \cnum{48.64} & \cnum{50.18} & \cnum{45.17} & \cnum{41.35} \\
\cmod{} & \cstg{After} & \cnum{80.50} & \cnum{79.91} & \cnum{79.90} & \cnum{79.93} & \cnum{80.14} \\
\midrule
\multicolumn{7}{l}{\textit{Concepts via Prompting}} \\
\midrule
\multirow{2}{*}{\cmod{Llama 1B}} & \cstg{Before} & \cnum{53.27} & \cnum{53.26} & \cnum{54.46} & \cnum{51.49} & \cnum{45.91} \\
\cmod{} & \cstg{After} & \cnum{80.44} & \cnum{80.47} & \cnum{80.42} & \cnum{80.60} & \cnum{80.61} \\
\dashsep{7}
\multirow{2}{*}{\cmod{Llama 3B}} & \cstg{Before} & \cnum{53.13} & \cnum{53.24} & \cnum{54.43} & \cnum{50.54} & \cnum{46.62} \\
\cmod{} & \cstg{After} & \cnum{81.42} & \cnum{81.43} & \cnum{81.30} & \cnum{81.28} & \cnum{81.18} \\
\dashsep{7}
\multirow{2}{*}{\cmod{Llama 8B}} & \cstg{Before} & \cnum{52.54} & \cnum{51.89} & \cnum{53.23} & \cnum{49.44} & \cnum{45.01} \\
\cmod{} & \cstg{After} & \cnum{82.41} & \cnum{82.20} & \cnum{82.27} & \cnum{82.18} & \cnum{82.49} \\
\dashsep{7}
\multirow{2}{*}{\cmod{Qwen 1.7B}} & \cstg{Before} & \cnum{42.50} & \cnum{44.80} & \cnum{46.07} & \cnum{40.28} & \cnum{34.57} \\
\cmod{} & \cstg{After} & \cnum{79.64} & \cnum{78.50} & \cnum{79.99} & \cnum{80.23} & \cnum{80.11} \\
\dashsep{7}
\multirow{2}{*}{\cmod{Qwen 4B}} & \cstg{Before} & \cnum{45.42} & \cnum{45.10} & \cnum{46.59} & \cnum{39.43} & \cnum{33.56} \\
\cmod{} & \cstg{After} & \cnum{81.12} & \cnum{80.93} & \cnum{80.94} & \cnum{80.84} & \cnum{81.20} \\
\dashsep{7}
\multirow{2}{*}{\cmod{Qwen 8B}} & \cstg{Before} & \cnum{41.64} & \cnum{30.34} & \cnum{42.12} & \cnum{29.35} & \cnum{22.21} \\
\cmod{} & \cstg{After} & \cnum{82.15} & \cnum{80.76} & \cnum{81.37} & \cnum{80.95} & \cnum{80.92} \\
\dashsep{7}
\multirow{2}{*}{\cmod{Falcon 1B}} & \cstg{Before} & \cnum{53.28} & \cnum{53.14} & \cnum{53.84} & \cnum{53.03} & \cnum{46.28} \\
\cmod{} & \cstg{After} & \cnum{75.83} & \cnum{75.14} & \cnum{75.17} & \cnum{75.00} & \cnum{75.10} \\
\dashsep{7}
\multirow{2}{*}{\cmod{Falcon 3B}} & \cstg{Before} & \cnum{53.89} & \cnum{53.11} & \cnum{54.56} & \cnum{51.62} & \cnum{44.06} \\
\cmod{} & \cstg{After} & \cnum{77.40} & \cnum{76.88} & \cnum{76.98} & \cnum{76.95} & \cnum{76.57} \\
\dashsep{7}
\multirow{2}{*}{\cmod{Falcon 7B}} & \cstg{Before} & \cnum{48.60} & \cnum{48.66} & \cnum{50.87} & \cnum{44.69} & \cnum{40.98} \\
\cmod{} & \cstg{After} & \cnum{80.44} & \cnum{79.90} & \cnum{80.19} & \cnum{79.92} & \cnum{80.05} \\
\bottomrule
\end{tabular}
\caption{STS (9 tasks) scores ($\times 100$) for every OpenWebText-trained model, before and after identical SimCSE-style contrastive finetuning. \textbf{Before} rows are the post-trained models the rest of the paper evaluates; \textbf{After} rows are those same models after contrastive finetuning. Before finetuning the five types within each model differ by as much as 20.4 points within a single row; afterwards they agree to within 1.73 points everywhere, including the randomized-synonym variants trained on deliberately corrupted supervision.}
\label{tab:contrastive_sts_owt}
\end{table*}

\begin{table*}[h!]
\centering
\footnotesize
\begin{tabular}{llccccc}
\toprule
\cmod{\textbf{Model}} & \cstg{} & \cnum{\textbf{Pretrained}} & \cnum{\textbf{NTP-only}} & \cnum{\textbf{Concept}} & \cnum{\textbf{\makecell{Rand.\\$\lambda{=}0.25$}}} & \cnum{\textbf{\makecell{Rand.\\$\lambda{=}1$}}} \\
\midrule
\multicolumn{7}{l}{\textit{Concepts via Embedding}} \\
\midrule
\multirow{2}{*}{\cmod{Llama 1B}} & \cstg{Before} & \cnum{47.08} & \cnum{47.15} & \cnum{48.27} & \cnum{45.26} & \cnum{41.52} \\
\cmod{} & \cstg{After} & \cnum{64.05} & \cnum{63.90} & \cnum{64.28} & \cnum{63.99} & \cnum{63.81} \\
\dashsep{7}
\multirow{2}{*}{\cmod{Llama 3B}} & \cstg{Before} & \cnum{48.34} & \cnum{47.84} & \cnum{48.80} & \cnum{44.71} & \cnum{41.12} \\
\cmod{} & \cstg{After} & \cnum{64.22} & \cnum{64.78} & \cnum{64.38} & \cnum{64.69} & \cnum{64.99} \\
\dashsep{7}
\multirow{2}{*}{\cmod{Llama 8B}} & \cstg{Before} & \cnum{49.10} & \cnum{48.40} & \cnum{48.98} & \cnum{44.06} & \cnum{40.25} \\
\cmod{} & \cstg{After} & \cnum{65.33} & \cnum{65.66} & \cnum{65.70} & \cnum{65.33} & \cnum{65.55} \\
\dashsep{7}
\multirow{2}{*}{\cmod{Qwen 1.7B}} & \cstg{Before} & \cnum{37.28} & \cnum{38.92} & \cnum{40.23} & \cnum{36.24} & \cnum{33.14} \\
\cmod{} & \cstg{After} & \cnum{64.38} & \cnum{64.34} & \cnum{64.41} & \cnum{64.72} & \cnum{64.71} \\
\dashsep{7}
\multirow{2}{*}{\cmod{Qwen 4B}} & \cstg{Before} & \cnum{39.29} & \cnum{39.07} & \cnum{41.08} & \cnum{36.03} & \cnum{32.94} \\
\cmod{} & \cstg{After} & \cnum{65.10} & \cnum{64.85} & \cnum{65.10} & \cnum{65.01} & \cnum{65.24} \\
\dashsep{7}
\multirow{2}{*}{\cmod{Qwen 8B}} & \cstg{Before} & \cnum{35.06} & \cnum{30.83} & \cnum{36.75} & \cnum{29.66} & \cnum{26.57} \\
\cmod{} & \cstg{After} & \cnum{65.69} & \cnum{65.55} & \cnum{65.36} & \cnum{65.68} & \cnum{65.60} \\
\dashsep{7}
\multirow{2}{*}{\cmod{Falcon 1B}} & \cstg{Before} & \cnum{47.40} & \cnum{47.77} & \cnum{49.12} & \cnum{47.29} & \cnum{42.02} \\
\cmod{} & \cstg{After} & \cnum{60.65} & \cnum{60.29} & \cnum{60.13} & \cnum{60.45} & \cnum{60.33} \\
\dashsep{7}
\multirow{2}{*}{\cmod{Falcon 3B}} & \cstg{Before} & \cnum{45.86} & \cnum{47.06} & \cnum{47.88} & \cnum{46.16} & \cnum{37.90} \\
\cmod{} & \cstg{After} & \cnum{62.11} & \cnum{61.73} & \cnum{61.72} & \cnum{61.73} & \cnum{62.03} \\
\dashsep{7}
\multirow{2}{*}{\cmod{Falcon 7B}} & \cstg{Before} & \cnum{43.40} & \cnum{43.10} & \cnum{44.55} & \cnum{40.55} & \cnum{37.87} \\
\cmod{} & \cstg{After} & \cnum{64.10} & \cnum{63.98} & \cnum{63.71} & \cnum{64.08} & \cnum{64.00} \\
\midrule
\multicolumn{7}{l}{\textit{Concepts via Prompting}} \\
\midrule
\multirow{2}{*}{\cmod{Llama 1B}} & \cstg{Before} & \cnum{47.08} & \cnum{47.20} & \cnum{48.25} & \cnum{45.73} & \cnum{41.99} \\
\cmod{} & \cstg{After} & \cnum{64.05} & \cnum{63.77} & \cnum{64.01} & \cnum{64.06} & \cnum{63.65} \\
\dashsep{7}
\multirow{2}{*}{\cmod{Llama 3B}} & \cstg{Before} & \cnum{48.34} & \cnum{47.84} & \cnum{49.12} & \cnum{45.18} & \cnum{41.74} \\
\cmod{} & \cstg{After} & \cnum{64.22} & \cnum{64.78} & \cnum{64.36} & \cnum{64.77} & \cnum{64.83} \\
\dashsep{7}
\multirow{2}{*}{\cmod{Llama 8B}} & \cstg{Before} & \cnum{49.11} & \cnum{48.24} & \cnum{49.47} & \cnum{44.82} & \cnum{40.02} \\
\cmod{} & \cstg{After} & \cnum{65.33} & \cnum{65.75} & \cnum{65.78} & \cnum{65.54} & \cnum{65.62} \\
\dashsep{7}
\multirow{2}{*}{\cmod{Qwen 1.7B}} & \cstg{Before} & \cnum{37.28} & \cnum{38.83} & \cnum{39.74} & \cnum{35.66} & \cnum{32.59} \\
\cmod{} & \cstg{After} & \cnum{64.38} & \cnum{63.90} & \cnum{64.43} & \cnum{64.82} & \cnum{64.75} \\
\dashsep{7}
\multirow{2}{*}{\cmod{Qwen 4B}} & \cstg{Before} & \cnum{39.29} & \cnum{39.07} & \cnum{40.16} & \cnum{35.10} & \cnum{31.86} \\
\cmod{} & \cstg{After} & \cnum{65.10} & \cnum{64.85} & \cnum{65.32} & \cnum{65.00} & \cnum{65.09} \\
\dashsep{7}
\multirow{2}{*}{\cmod{Qwen 8B}} & \cstg{Before} & \cnum{35.06} & \cnum{30.86} & \cnum{35.64} & \cnum{29.29} & \cnum{26.96} \\
\cmod{} & \cstg{After} & \cnum{65.39} & \cnum{65.81} & \cnum{65.59} & \cnum{66.08} & \cnum{65.53} \\
\dashsep{7}
\multirow{2}{*}{\cmod{Falcon 1B}} & \cstg{Before} & \cnum{47.36} & \cnum{47.90} & \cnum{47.79} & \cnum{47.12} & \cnum{40.63} \\
\cmod{} & \cstg{After} & \cnum{60.65} & \cnum{60.22} & \cnum{60.26} & \cnum{60.37} & \cnum{60.09} \\
\dashsep{7}
\multirow{2}{*}{\cmod{Falcon 3B}} & \cstg{Before} & \cnum{45.86} & \cnum{46.85} & \cnum{47.43} & \cnum{45.20} & \cnum{37.56} \\
\cmod{} & \cstg{After} & \cnum{62.11} & \cnum{61.93} & \cnum{61.78} & \cnum{61.96} & \cnum{61.83} \\
\dashsep{7}
\multirow{2}{*}{\cmod{Falcon 7B}} & \cstg{Before} & \cnum{43.42} & \cnum{42.93} & \cnum{44.71} & \cnum{40.08} & \cnum{37.37} \\
\cmod{} & \cstg{After} & \cnum{64.03} & \cnum{64.15} & \cnum{64.09} & \cnum{64.02} & \cnum{64.01} \\
\bottomrule
\end{tabular}
\caption{Short-context MTEB (15 tasks) scores ($\times 100$) for every OpenWebText-trained model, before and after identical SimCSE-style contrastive finetuning. \textbf{Before} rows are the post-trained models the rest of the paper evaluates; \textbf{After} rows are those same models after contrastive finetuning. Before finetuning the five types within each model differ by as much as 10.2 points within a single row; afterwards they agree to within 0.92 points everywhere, including the randomized-synonym variants trained on deliberately corrupted supervision.}
\label{tab:contrastive_short_owt}
\end{table*}

\end{document}